\pdfoutput=1
\documentclass{article} %
\usepackage{iclr2024_conference,times}

\usepackage{amsmath,amsfonts,bm}

\definecolor{tablecolor}{rgb}{0.8,0.8,0.8}

\newcommand{\vparam}{\vtheta}

\newcommand{\ptheta}{p_{\scaleto{\vparam}{4pt}}}
\newcommand{\pTheta}{p_{\scaleto{\Theta}{4pt}}}
\newcommand{\phatTheta}{\widehat{p}_{\scaleto{\Theta}{4pt}}}

\newcommand{\vocab}{\mathcal{V}}
\newcommand{\hypset}{\mathcal{H}}
\newcommand{\ensemble}{\mathcal{M}}

\newcommand\cut[1]{}

\newcommand{\squishlist}{
   \begin{list}{$\bullet$}
    { \setlength{\itemsep}{0pt}      \setlength{\parsep}{3pt}
      \setlength{\topsep}{3pt}       \setlength{\partopsep}{0pt}
      \setlength{\leftmargin}{1.5em} \setlength{\labelwidth}{1em}
      \setlength{\labelsep}{0.5em} } }

\newcommand{\squishlisttwo}{
   \begin{list}{$\bullet$}
    { \setlength{\itemsep}{0pt}    \setlength{\parsep}{0pt}
      \setlength{\topsep}{0pt}     \setlength{\partopsep}{0pt}
      \setlength{\leftmargin}{2em} \setlength{\labelwidth}{1.5em}
      \setlength{\labelsep}{0.5em} } }

\newcommand{\squishend}{
    \end{list}  }

{}
{}
{}

\newcommand{\real}{\mbox{$\mathbb{R}$}}

\newcommand{\myexpect}{\mathbb{E}}

\newcommand{\gauss}{\mbox{${\cal N}$}}

\newcommand{\myvec}[1]{\mathbf{#1}}
\newcommand{\myvecsym}[1]{\mbox{$\boldsymbol{#1}$}}

\newcommand{\vtheta}{\mathbf{\boldsymbol{\theta}}}

\newcommand{\vsigma}{\mbox{$\myvecsym{\sigma}$}}
\newcommand{\vSigma}{\mbox{$\myvecsym{\Sigma}$}}

\newcommand{\vh}{\mbox{$\myvec{h}$}}

\newcommand{\vm}{\mbox{$\myvec{m}$}}

\newcommand{\vx}{\mbox{$\myvec{x}$}}

\newcommand{\vy}{\myvec{y}}

\newcommand{\calD}{\mbox{${\cal D}$}}

\newcommand{\data}{\calD}

\usepackage{hyperref}
\usepackage{url}
\usepackage{times}
\usepackage{latexsym}
\usepackage{amsmath}
\usepackage{amssymb}
\usepackage{mathtools}
\usepackage{bbm}
\usepackage{cleveref}
\usepackage{footnote}

\crefname{section}{\S}{\S\S}
\Crefname{section}{\S}{\S\S}
\crefname{table}{Tab.}{}
\crefname{figure}{Fig.}{Figs.}
\crefname{algorithm}{Alg.}{}
\crefname{equation}{Eq.}{Eqs.}
\crefname{appendix}{App.}{}
\crefname{theorem}{Theorem}{}
\crefname{proposition}{Proposition}{}
\crefname{defin}{Definition}{}
\crefname{cor}{Corollary}{}
\crefname{observation}{Observation}{}
\crefname{assumption}{Assumption}{}
\crefformat{section}{\S#2#1#3}
\crefformat{footnote}{#2\footnotemark[#1]#3}

\usepackage{pgfplots}
\usepackage{subcaption}
\usepackage{tikz}
\usepackage{scalerel}

\DeclareMathOperator*{\argmax}{arg\,max}

\usepackage{microtype}

\usepackage{inconsolata}
\definecolor{brandeisblue}{rgb}{0.0, 0.44, 1.0}
\definecolor{brandeisbluecompl}{HTML}{FF8F00}
\definecolor{brandeis2}{HTML}{0F00FF}
\definecolor{brandeis4}{HTML}{00F0FF}
\definecolor{brandeispurple}{HTML}{6D00FF}
\definecolor{palette1}{HTML}{0070FF}
\definecolor{palette2}{HTML}{5943AA}
\definecolor{palette3}{HTML}{C65454}
\definecolor{palette4}{HTML}{E88262}

\newcommand{\revision}[1]{{\color{black}#1}}

\title{Uncertainty-Aware Decoding with \\Minimum Bayes Risk}

\author{
Nico Daheim$^{1}$, Clara Meister$^2$, Thomas M\"ollenhoff$^3$, Iryna Gurevych$^{1}$\\
    $^1$Ubiquitous Knowledge Processing Lab (UKP Lab)
\\ \;\,Department of Computer Science and Hessian Center for AI (hessian.AI)
\\ \;\,Technical University of Darmstadt \;\, $^2$ETH Zurich \\ $^3$RIKEN Center for Advanced Intelligence Project, Tokyo, Japan\\
\;\,\url{www.ukp.tu-darmstadt.de}
}

\begin{document}

\maketitle

\begin{abstract}

Despite their outstanding performance in the majority of scenarios, contemporary language models still occasionally \revision{generate} undesirable outputs, for example, hallucinated text. 
While such behaviors have previously been linked to uncertainty, there is a notable lack of methods that actively consider uncertainty during text generation.
In this work, we show how Minimum Bayes Risk (MBR) decoding,
\revision{which selects model generations according to an expected risk,}
can be generalized into a principled uncertainty-aware decoding method. 
In short, we account for model uncertainty during decoding by incorporating a posterior over model parameters into MBR’s computation of expected risk.  
We show that this modified expected risk is useful for both choosing outputs and deciding when to abstain from generation \revision{and can provide improvements without incurring overhead}.
We benchmark different methods for learning posteriors and show that performance \revision{improves with prediction diversity. We release our code publicly.\footnote{\url{https://github.com/UKPLab/iclr2025-mbr-uncertainty}}%
}
\end{abstract}

\section{Introduction}

Today's language models can generate fluent and coherent text. 
While they perform well in many scenarios, there are still instances where they fail and, for example, hallucinate factually incorrect outputs or generate harmful language \citep{ye2023cognitivemiragereviewhallucinations, bhandari-brennan-2023-trustworthiness, li-etal-2024-dawn}. 
\revision{Previous works have shown that these behaviors are often related to out-of-distribution inputs~\citep{ren2023outofdistribution} and (epistemic) uncertainty~\citep{xiao-wang-2021-hallucination, van-der-poel-etal-2022-mutual, fadeeva-etal-2024-fact} which are both connected to uncertainty about the parameters of the model.}
Yet there is still a lack of methods that adjust for this type of uncertainty during decoding in language generation.

Minimum Bayes Risk (MBR) decoding was originally proposed for statistical machine translation~\citep{kumar-byrne-2002-minimum}, motivated by similar model shortcomings.
The idea of MBR is to make use of the entire distribution when choosing an output, because, while the model distribution might be a good overall representation of the target distribution~\citep{smith:2011:synthesis}, individual samples might not be adequate.
More recent works have shown that such problems persist with modern models~\citep{stahlberg-byrne-2019-nmt,pmlr-v97-cohen19a, eikema-aziz-2020-map}, precipitating the resurgence of MBR~\citep{freitag-etal-2022-high}.
In this work, we show how a small adjustment to MBR decoding can enhance it beyond this scope and turn it into an uncertainty-aware decoding method.

In short, we modify MBR's definition of expected risk by incorporating an additional expectation over a posterior distribution over model parameters. 
This adjustment enables us to account for uncertainty in parameter estimates when judging the quality of different hypotheses from a model. 
\revision{We present different estimators for this expected risk which use multiple models from the (approximate) posterior to generate outputs.\footnote{This approach has previously been shown to improve downstream performance in classification tasks~\citep{blundell2015weight, lakshminarayanan2017simple, maddox2019simple, shen2024variational}.}
Two of these estimators combine outputs at the sequence-level, i.e. full strings generated by a model, which is useful for combining the outputs of black-box LLMs for which one does not have access to output probabilities. %
Another estimator combines token-level distributions.}

Overall, we find strong evidence that accounting for weight uncertainty can improve decoding and reduce hallucinations \revision{when finetuning and pretraining from scratch}, \revision{even without computational overhead}.
We find that improvements trend with the expressiveness of the posterior.
Likely related to this, the performance of uncertainty-aware MBR is highly correlated with the prediction diversity across the combined models.
We also find that weight uncertainty provides a useful signal for selective prediction, where we observe that \revision{the} uncertainty-aware expected risk can be used to decide when to predict or abstain from generation. 
Furthermore, we show that performance scales: it improves with more models and larger hypothesis set sizes.
Finally, we show the effectiveness of this framework when used to ensemble outputs from black-box LLMs.
\section{Background}
\subsection{Probabilistic Language Generation}
Modern models for language generation are predominantly locally-normalized, autoregressive models of a conditional distribution over next tokens.
The probability of a sequence of tokens forming a string can be determined by the product of all next-token probabilities in the sequence. 
Formally, given input $\vx$ and model $\ptheta$
the probability of an output sequence $\vy = \langle y_1, y_2, \dots \rangle$ is computed as
\begin{equation}
    \ptheta(\vy\mid \vx) = \prod_{t=1}^{|\vy|}\ptheta(y_t\mid \vy_{<t}, \vx).
\end{equation}
Here, each $y_t$ is a token from some predetermined vocabulary $\vocab$ and $\vparam \in \real^d$ are the parameters of the model which are also often called weights. The input $\vx$ could be text but, for example, also images.

\paragraph{Learning $\smash{\ptheta}$.} 
The parameters of the model $\smash{\ptheta}$ are generally learned given paired examples $\smash{\data = \{\vx^{(i)}, \vy^{(i)}\}_{i=1}^N}$, a loss function and an optimization procedure.
The loss function then indicates how well the model $\smash{\ptheta}$ captures the data-generating distribution $\smash{p(\cdot\mid\vx)}$ \revision{from which we assume $\smash{\data}$ is sampled.}
In most cases, language generation models are learned by minimizing an empirical risk over data examples in terms of one parameter set $\smash{\vparam \in \real^d}$, for example, using AdamW~\citep{loshchilov2018decoupled}.
However, such approaches can not directly model weight uncertainty. 
In this work, we instead use Bayesian methods to model weight uncertainty. We describe them in~\cref{subsec:generalizing_mbr} and \cref{sec:learning_weight_uncertainty}.

\paragraph{Decoding from $\smash{\ptheta}$.} At inference time, our goal is to generate a string from $\smash{\ptheta(\cdot\mid\vx)}$. 
The set of decision rules used in this process is often referred to as the decoding strategy. 
One such strategy is simply to sample tokens autoregressively until a stopping criterion, usually a fixed maximum length or a special end-of-sequence token, is met. 
Another strategy is to {(approximately)} search for the maximum probability string according to $\smash{\ptheta(\cdot\mid\vx)}$. %
Both of these approaches have proved problematic empirically \citep{fan-etal-2018-hierarchical, Holtzman2020The, eikema-aziz-2020-map, hewitt-etal-2022-truncation}, prompting the exploration of alternative strategies. 
The shortcomings of { all of} these strategies have been (at least partially) attributed to the fact that they do not consider a string's utility, which may not perfectly align with its probability. 
Minimum Bayes Risk decoding aims to solve this issue. 

\subsection{Minimum Bayes Risk Decoding}
Minimum Bayes Risk decoding is derived from Bayesian Decision Theory, which states that optimal decisions are those that minimize an expected risk or, equivalently, maximize an expected utility \citep[see][inter alia]{degroot2005optimal}. 
Given a utility function $u: \smash{\vocab^\ast \times \vocab^\ast \rightarrow \mathbb{R}_{\geq 0}}$ 
which assigns to each pair of strings a non-negative utility, MBR aims to find the string that maximizes expected utility with respect to the target distribution. 
This principle is especially appealing when working with a possibly imperfect model of the target distribution, such as $\ptheta$, because it allows using the full model distribution instead of relying on the adequacy of individual samples, which is argued to be the downfall of other decoding strategies \citep{eikema-aziz-2020-map}. 
We thus choose the hypothesis:
\begin{align}
    \vy^\ast &= \argmax_{\vy^\prime \in \vocab^\ast} \underset{\vy \sim \ptheta(\cdot\mid\mathbf{x})}{{\myexpect}}\left[ u(\vy, \vy^\prime)\right] \label{eq:expect} \\
    &= \argmax_{\vy^\prime \in \vocab^\ast} \sum_{\vy \in \vocab^\ast}\ptheta(\vy\mid\mathbf{x}) u(\vy, \vy^\prime). \label{eq:sum}
\end{align}
There are several obstacles to computing \cref{eq:sum}. 
Both summing over all possible strings in $\vocab^\ast$ to compute {the} expectation and searching over them to find the expectation-maximizing hypothesis are computationally infeasible.\footnote{The latter problem is not unique to MBR, and faced by all maximization-based decoding strategies for autoregressive language generators. Hence, approximation algorithms are also used for these strategies. }
Thus, {approximations to~\cref{eq:sum} are used in practice.}
 
The common approach to circumvent these obstacles is to employ an {(often Monte Carlo)} estimator of the expected utility and limit the search space to a subset of $ \smash{\vocab^\ast}$. 
Since the estimator requires a sample of strings from the distribution of interest, the same strings are often used in both the utility estimation and approximate search.\footnote{Some works have explored using different subsets for these two steps~\citep{eikema-aziz-2022-sampling, fernandes-etal-2022-quality}; we leave the exploration of the interaction of this design choice with our methods to future work.} We refer to this collection as the hypothesis set and denote the samples used in our estimator as $\smash{\hypset = [\vy^{(i)}]_{i=1}^N}$. In the case of a Monte Carlo estimator, where all $\smash{\vy^{(i)}\sim\ptheta}$, we denote this collection as $\hypset_\vparam$.  
This leads to the following approximation to \cref{eq:sum}:\footnote{We drop the normalizing term for succinctness as it does not affect the $\argmax$ operation. \label{mc_footnote}}
\begin{equation}
    \widehat\vy^\ast = \argmax_{\vy^\prime \in \hypset_\vparam} \sum_{\vy \in \hypset_\vparam} u(\vy, \vy^\prime), \label{eq:approx}
\end{equation}
\revision{which has proven to be a useful criterion for selecting examples for knowledge distillation~\citep{finkelstein2024mbr, yang-etal-2024-direct}}.
Most prior work \revision{on MBR} has focused on making the approximation in \cref{eq:approx} more efficient \citep{eikema-aziz-2022-sampling, fernandes-etal-2022-quality, cheng-vlachos-2023-faster, vamvas-sennrich-2024-linear} or on better choices for utility functions \citep{freitag-etal-2022-high} but few have considered an important underlying assumption: that $\ptheta$ is a good substitute for $p$. 
If uncertainty over the suitable model parameters $\vparam$ (i.e. weight uncertainty)
is high, e.g., when training data is limited, using a single $\ptheta$ may not provide a good substitute.
Bayesian modeling already provides tools to account for such uncertainty by marginalizing a distribution over possible parameters. 
We use this approach next to establish uncertainty-aware decoding schemes that account for weight uncertainty.

\section{Minimum Bayes Risk Decoding with Weight-Uncertainty}
In this section, we show how a simple change can turn MBR into an uncertainty-aware decoding method.
We first introduce weight uncertainty.
Then, we use it to establish an uncertainty-aware variant of MBR before presenting three practical decoding methods based on it.

\subsection{Generalizing MBR with Weight Uncertainty}
\label{subsec:generalizing_mbr}
\revision{Often, a distribution over possible model parameters is used to model weight uncertainty based on Bayesian principles~\citep{maddox2019simple, osawa2019practical, mollenhoff2023sam}.}
\revision{According to Bayes' theorem, the probability of a parameterization $\vparam$ is} $\smash{p(\vparam\mid\data) \propto p(\data\mid\vparam)\cdot p(\vparam)}$ where $\smash{p(\vparam)}$ is a prior and $\smash{\data}$ is our data.
\revision{Because} calculating an exact distribution $\smash{p(\vparam\mid\data)}$ over model parameters is intractable, an approximate distribution $\smash{q(\cdot)}$ is usually used.
There are numerous methods one can use for obtaining $\smash{q(\cdot)}$, \revision{for example, Laplace~\citep{mackay1992bayesian, laplace2021} and variational learning~\citep{graves2011practical, blundell2015weight}.
We use variational learning, which we describe in~\cref{sec:learning_weight_uncertainty}}.

Access to a posterior $\smash{q(\cdot)}$ allows prediction by combining the outputs of multiple $\smash{\ptheta}$, weighted by the probability $\smash{q(\vparam)}$ of each parameterization $\smash{\vparam}$. 
The resulting distribution is often referred to as the {predictive posterior} distribution, which we denote as $\smash{\pTheta}$.
Empirically, this has been shown to improve calibration~\citep{yang2024bayesian} and uncertainty estimation~\citep{shen2024variational}.
However, in modern language generation, it is not immediately clear how model predictions should be combined in practice.
Combining predictions in probability space is difficult, \revision{because many modern LLMs are only accessible via APIs that do not return probabilities at either the token- or sequence-level.}
Standard Monte-Carlo-based methods avoid this issue, but they \revision{may be} problematic: even for larger sample sizes, a given string would likely only be sampled once. 
While generations might be approximately similar, e.g., differing only in punctuation, this approach treats them as completely disparate. 
We now show how a natural extension of MBR provides a \revision{principled} framework for combining model predictions that circumvents these issues.

{%
\revision{We propose to generalize MBR by replacing the definition of $\ptheta$ in~\cref{eq:sum} with the predictive posterior $\pTheta$ to account for weight uncertainty. Then, the search problem is}:\footnote{\revision{Here, $\Theta$ denotes all possible parameterizations $\vparam$ of the model and is used to indicate a predictive posterior.}}} 
\begin{equation}
    \vy^{\Theta} = \argmax_{\vy^\prime\in\vocab^\ast} \sum_{\vy\in\vocab^\ast} \pTheta(\vy\mid\vx) u(\vy, \vy^\prime). 
\label{eq:uncertainty_mbr}
\end{equation} 
We recover standard MBR when using the delta method to approximate $\pTheta$, i.e., approximating the predictive posterior using one model parameterized by the mean of $q$~\citep[App. C]{khan2023bayesian}. Monte-Carlo-based approximations of MBR do not require knowledge of string probabilities but only the ability to sample from the model. Further, the utility function can be chosen as a soft matching between strings to account for similarities between samples instead of treating them as completely distinct, which addresses the aforementioned issues.
\revision{We next discuss three decoding algorithms for approximately solving~\cref{eq:uncertainty_mbr} using sequence- (\cref{subsec:seq_level_mbr}) and token-level posteriors (\cref{subsec:tok_level_mbr}).\footnote{\revision{We refer to~\citet[Sec. 3, App.A]{malinin2021uncertainty} for further discussion of token- and sequence-level posteriors for uncertainty estimation for autoregressive models.}}}

\subsection{Sequence-Level Posteriors for Uncertainty-Aware Decoding}
\label{subsec:seq_level_mbr}
While autoregressive language models are trained to model a distribution over tokens, the quantity of interest is often the probability of an entire sequence \revision{$\smash{\vy \in \vocab^\ast}$}. Therefore, it is natural to model a predictive posterior on a sequence-level by using an expectation over sequence probabilities: \begin{equation}
    \pTheta^{\text{(seq)}}(\vy \mid \vx) \coloneqq  \underset{\vparam\sim q}{\myexpect} \left[\ptheta(\vy \mid \vx)\right ] = \underset{\vparam\sim q}{\myexpect} \left[\prod_{t=1}^{|\vy|}\ptheta(y_t\mid \vy_{<t}, \vx)\right]
    \label{eq:seq_avg}
\end{equation}
Using \revision{a sequence-level} posterior to replace the model distribution in~\cref{eq:sum} \revision{admits two practical methods for (soft) model averaging, because} when $u$ is bounded or non-negative\footnote{Many commonly used utility functions for MBR are bounded and non-negative.
For example, BLEU~\citep{papineni-etal-2002-bleu} and BERTScore~\citep{Zhang2020BERTScore} return scores from 0 to 100 or 0 to 1, respectively.} \revision{the order of the two expectations in~\cref{eq:uncertainty_mbr} can be switched due to Fubini's theorem \citetext{\citealp[Sec. 8.9]{degroot2005optimal}; \citealp[Sec. 2.3]{robert2007bayesian}}. 
Note that the second expectation (over models) is implicit in the definition of $\smash{\pTheta^{\text{(seq)}}}$. 
This suggests using the equivalence}: 
\begin{align}
    \vy^\Theta &= \argmax_{\vy^\prime\in\vocab^\ast} \sum_{\vy\in\vocab^\ast} \underset{\vparam\sim q}{\myexpect}\left[\ptheta(\vy\mid\vx)\right ]u(\vy, \vy^\prime) \label{eq:insert_def}\\
    &= \argmax_{\vy^\prime\in\vocab^\ast} \underset{\vparam\sim q}{\myexpect} \Big[\sum_{\vy\in\vocab^\ast}\ptheta(\vy\mid\vx)u(\vy, \vy^\prime)\Big].  
    \label{eq:seq_level_mbr}
\end{align}

\revision{Monte-Carlo estimators of~\cref{eq:insert_def} and~\cref{eq:seq_level_mbr} then correspond to either 
sampling a collection of generations from a set of models $\smash{\ensemble = \{\vparam^{(i)} \sim q(\vparam) \}_{i=1}^M}$ sampled i.i.d from $q$ and using~\cref{eq:approx} or using~\cref{eq:approx} independently for each model before summing the utilities of each output across models.
Formally, let $\hypset_{\vparam}$ denote the hypothesis set for each model $\vparam$ and $\hypset_\ensemble = \uplus_{\vparam\in\ensemble} \hypset_{\vparam}$ the collection of all generated outputs across them.
Note that both $\hypset_\ensemble$, indicated by $\uplus$, and $\hypset_\vparam$ preserve sample counts \revision{to ensure an unbiased estimator}. 
Then, the} approximate solutions become:\cref{mc_footnote}

\begin{minipage}{.5\linewidth}    
\begin{equation}
    \widehat\vy^\Theta 
    = \argmax_{\vy^\prime\in \hypset_\ensemble } \sum_{\vy\in\hypset_{\ensemble}}u(\vy, \vy^\prime)
    \label{eq:hyp_set_concat_estimator}
\end{equation}
\end{minipage}%
\begin{minipage}{.5\linewidth}
    \begin{equation}
         \widehat\vy^\Theta = \argmax_{\vy^\prime\in \hypset_\ensemble } \sum_{\vparam \in \ensemble} \sum_{\vy\in\hypset_{\vparam}}u(\vy, \vy^\prime). 
    \label{eq:seq_level_mbr_estimator}
    \end{equation}
\end{minipage}

This is convenient because it allows us to ensemble any set of LLMs given just the ability to sample from them and can easily be parallelized. \revision{No access to probabilities is required.}
For \cref{eq:seq_level_mbr_estimator}, even utility computation can be parallelized \revision{across models.
There are trade-offs between both estimators, especially in terms of computational complexity, which we discuss next.}

\paragraph{\revision{Computational Costs.}}\cref{eq:hyp_set_concat_estimator} requires $\smash{(|\ensemble|\cdot|\hypset_{\vparam}|)^2}$ \revision{evaluations of $u$}. This might be impractical for large sizes of $\smash{\hypset_{\vparam}}$ but, intuitively, the larger amount of comparisons might be helpful for MBR.
\cref{eq:seq_level_mbr_estimator} requires only $\smash{|\ensemble|\cdot|\hypset_{\vparam}|^2}$ utility computations. \revision{This is faster and, especially for more costly utility functions that e.g. use LLMs as judges~\citep{wu2025better}, can enable larger hypothesis set sizes.} 

\paragraph{\revision{Discussion.}}
\revision{When preserving sample counts,~\cref{eq:hyp_set_concat_estimator} provides an unbiased estimate of~\cref{eq:insert_def}. Intuitively, this is advantageous because highly probable sequences can contribute more to the decision.}
This differentiates ours from prior work, such as~\citet[Alg. 1]{kobayashi-2018-frustratingly}, who rather use a set union. 
Recent work~\citep{farinhas-etal-2023-empirical} also uses~\cref{eq:hyp_set_concat_estimator} but does not explore the connection to weight uncertainty.
Our methods draw parallels between MBR, which aims to minimize expected risk, and PAC-Bayes bounds~\citep{alquier2021user}, which study the expected risk of predictive posteriors.
Finally, it also helps to understand early system aggregation methods that use similar decision rules as shown here, e.g., by optimizing scalar model weights~\citep[Eq. 8]{gonzalez-rubio-etal-2011-minimum}.

\subsection{Token-Level Posteriors for Uncertainty-Aware Decoding}
\label{subsec:tok_level_mbr}
\revision{While our definition of uncertainty-aware MBR uses sequence-level Bayesian modeling, language models generally define distributions over tokens. It is thus natural to consider a predictive posterior defined over token-level distributions, i.e.,  by averaging token-level probabilities at each time-step}: \begin{equation}
    \pTheta^{\text{(tok)}}(\vy \mid \vx) \coloneqq  \prod_{t=1}^T \underset{\vparam\sim q}{\myexpect} \left[\ptheta(y_t \mid \vy_{<t}, \vx)\right ].
    \label{eq:token_avg}
\end{equation}
\revision{Note, though, that for a given sequence, this approach will in general not assign the same probabilities as sequence-level ensembling.\footnote{This is because sequence-level modeling uses a an \emph{expectation of products} approach while token-level modeling uses a \emph{product of expectations} approach. Since expectation and product operations do not necessarily commute, these two ensemble definitions will, in general, assign different probabilities to the same sequence~\citep{malinin2021uncertainty}.}
Consequently, the decisions when using MBR might also be different.} Since \revision{the expectation over models is intractable in~\cref{eq:token_avg}}, we use a Monte Carlo estimator. \revision{The estimator averages the token-level probabilities given by the models $\smash{\ensemble}$ during generation}:  %
\begin{equation}
    \phatTheta^{\text{ (tok)}}(y_t\mid\vy_{<t},\vx) = \frac{1}{|\ensemble|}\sum_{\vparam \in \ensemble} \ptheta(y_t\mid\vy_{<t},\vx).
\end{equation}
When sampling the hypotheses set $\hypset_{\Theta}$ from this distribution, i.e., sampling each token according to $\smash{\phatTheta^{\text{ (tok)}}}$, an MBR estimator like the one in \cref{eq:uncertainty_mbr} can be used to incorporate weight-uncertainty:\cref{mc_footnote}
\begin{equation}
    \widehat\vy^\Theta 
    = \argmax_{\vy^\prime\in \hypset_{\Theta} }  \sum_{\vy\in\hypset_{\Theta}}u(\vy, \vy^\prime).\label{eq:tok_level_mbr}
\end{equation}

There are several intuitive reasons why this
\revision{could} improve decoding.
Perhaps the foremost is that probabilities obtained from model averaging are {often} better-calibrated than those of a single model~\citep[inter alia]{yang2024bayesian, shen2024variational}.
\revision{By averaging predictions, modeling weight uncertainty (over $\vparam$) can enable better estimates of predictive uncertainty (over $\vy$).}
Since predictive uncertainty has been shown to correlate with hallucinations~\citep{xiao-wang-2021-hallucination}, one hope would be that \revision{better estimates of it} would downweigh e.g. hallucinated outputs.

{\paragraph{\revision{Computational Costs.}} Token-level posteriors only require $\smash{|\hypset|^2}$-many MBR comparisons when the hypothesis set size is equal to $\smash{|\hypset|}$. Sequence-level combination requires $\smash{|\ensemble|\cdot |\hypset|^2}$-many comparisons for~\cref{eq:seq_level_mbr_estimator} or even $\smash{(|\ensemble|\cdot |\hypset|)^2}$-many comparisons for~\cref{eq:hyp_set_concat_estimator} if all hypothesis sets have the same size.
However, fitting all models for token-level combination on one GPU can be hard and communication overhead is high when distributing them across GPUs.
Further, token-level posteriors can not be used with black-box APIs that do not provide token-level probabilities.}

\subsection{Selective Prediction with Bayes Risk}
For some inputs{, for example, grammatically-incorrect strings, even a good model may not provide good predictions. 
Then, it can be wise} to abstain from answering and, e.g., defer to a human expert instead. 
{Selective prediction tackles this by abstaining} for inputs (or outputs) that score highly {in} some criterion $\smash{s: \vocab^{\ast} \rightarrow \real}$ that assigns a score for a given input $\vx$.~\citep{geifman2017selective, ren2023outofdistribution, kuhn2023semantic}. 
In practice, given $\smash{\alpha > 0}$ and a test dataset $\smash{\data_{\text{test}}}$, we only evaluate the model's answers for the top-$\smash{\lceil\alpha\cdot|\data_{\text{test}}|\rceil}$ examples according to $s$. 
If $s$ is reliable, performance should improve as $\alpha$ decreases and we evaluate a smaller and smaller subset of outputs.  

Expected utility {promises to be a good} criterion: 
if {we expect low utility}, we should abstain from answering; if {we expect high utility}, we can place more trust in the model's answer. 
We compare different methods for using expected utility as the selective prediction criterion.
We first consider the maximum-utility output in $\hypset_\Theta$ or $\hypset_\ensemble$ for~\cref{eq:tok_level_mbr} and~\cref{eq:seq_level_mbr_estimator}, i.e:\cref{mc_footnote}
\begin{align}
    s^\ast_{\text{tok}}(\vx) &=  \max_{\vy^\prime \in\hypset_\Theta} \sum_{\vy \in\hypset_\Theta} u(\vy, \vy^\prime) \quad \quad s^\ast_{\text{seq}}(\vx) = \max_{\vy^\prime \in \hypset_\ensemble} \sum_{\vparam\in\ensemble}\sum_{\vy\in\hypset_{\vparam}}u(\vy, \vy^\prime).
\end{align}
Note that we can easily define a similar risk for~\cref{eq:hyp_set_concat_estimator} by replacing $\hypset_\Theta$ with $\hypset_\ensemble$ in the definition of $\smash{s^\ast_{\text{tok}}(\vx)}$.
Another strategy is to use the expected utility \emph{across} outputs for the given input. 
We can do this by averaging the utility of all outputs in the hypothesis set $\hypset_\Theta$ or $\hypset_\ensemble$.\cref{mc_footnote}
\begin{align}
    \bar{s}_{\text{tok}}(\vx) &=  \sum_{\vy^\prime \in\hypset_\Theta} \sum_{\vy \in\hypset_\Theta} u(\vy, \vy^\prime) \quad \quad \bar{s}_{\text{seq}}(\vx) = \sum_{\vparam\in\ensemble}\sum_{\vy^\prime \in\hypset_{\vparam}}\sum_{\vy\in\hypset_{\vparam}}u(\vy, \vy^\prime).
\end{align}

\section{Experiments \& Results}
Here, we demonstrate empirically that incorporating weight uncertainty can improve decoding.
{First, we provide brief experimental details and discuss how we learn weight uncertainty in~\cref{sec:exp_details} \revision{and~\cref{sec:learning_weight_uncertainty}}. More details about our experiments are found in~\cref{app:experimental_details}.
Then, we show results using prompted, finetuned and from-scratch-trained models in~\cref{sec:experiments_main}, where we explore different posteriors and model combination methods.}
\Cref{sec:quality_diversity} looks into the trade-off between performance and ensemble diversity and~\cref{sec:selective_prediction} 
Bayes risk for selective prediction.
Finally, we {show} the scaling behavior of various methods in~\cref{sec:scaling}.\looseness=-1

\subsection{Experimental Details}
\label{sec:exp_details}

{
\paragraph{Datasets.}
We use WMT14~\citep{bojar14wmt}, IWSLT14~\citep{cettolo2014iwslt}, afroMT~\citep{reid-etal-2021-afromt}, IWSLT17~\citep{cettolo-etal-2017-overview}, WMT18~\citep{bojar-etal-2018-findings}, and WMT19~\citep{barrault-etal-2019-findings} for machine translation, XSUM~\citep{narayan-etal-2018-dont} and SAMSum~\citep{gliwa-etal-2019-samsum} for summarization, E2E-NLG~\citep{novikova2017e2e} for data-to-text generation, and STS-B~\citep{cer-etal-2017-semeval} for scoring.
For the latter, the model outputs a string representation of its numerical prediction and MBR corresponds to an empirical mean of the numerical predictions~\citep{lukasik2024metricawarellminferenceregression}.

\paragraph{Models.}
We zero-shot prompt Llama-3 8B~\citep{dubey2024llama3herdmodels}, Mistral 7B~\citep{jiang2023mistral7b}, Gemma-2 9B~\citep{gemmateam2024gemma2improvingopen}, and Qwen-2 7B~\citep{yang2024qwen2technicalreport}.
We finetune Gemma-2B-it~\citep{gemmateam2024gemmaopenmodelsbased} using LoRA~\citep{hu2022lora} with ca. 0.9M trainable parameters.
For training from scratch, we use the Transformer$_\text{big}$ architecture with ca. 261M parameters for WMT14 and Transformer$_\text{base}$ with 86M-126M parameters otherwise, following~\citet{vaswani2017}.

\paragraph{Metrics.}
For machine translation, we use the SacreBLEU implementation~\citep{post-2018-call} of BLEU~\citep{papineni-etal-2002-bleu}, chrF~\citep{popovic-2015-chrf}, the quality estimator COMET$_{22}$~\citep{rei-etal-2022-comet}, and LaBSE~\citep{feng-etal-2022-language} to evaluate hallucinations which has shown strong correlation with human judgements~\citep{dale-etal-2023-detecting,himmi2024enhanced}.
For Summarization and data-to-text generation we use ROUGE~\citep{lin-2004-rouge} and regression is evaluated using root mean-squared error (RMSE).
We use FactCC for hallucination evaluation on XSUM~\citep{kryscinski-etal-2020-evaluating}.
For the utility function $u$ we use BERTScore~\citep{Zhang2020BERTScore}, except for IWSLT14 and afroMT, where we use BLEU.
}

\subsection{\revision{Learning weight uncertainty.}}
\label{sec:learning_weight_uncertainty}
We use the variational learning algorithm IVON~\citep{shen2024variational} to estimate a posterior distribution over model weights \revision{and model weight uncertainty. We choose it, because each training run with IVON has only negligible overhead compared to AdamW~\citep{loshchilov2018decoupled} and gives comparable performance, as also shown in~\cref{tab:llm}.}
\revision{It is also possible to use other Bayesian Deep Learning methods, such as, Laplace~\citep{laplace2021} or SWAG~\citep{maddox2019simple} but we leave their exploration for future work.}
{IVON learns a unimodal Gaussian posterior $q(\vparam) \coloneqq \gauss(\vparam\mid\vm, \vSigma)$ with mean $\vm$ and \revision{(diagonal)} covariance matrix $\vSigma$.}
Setting model parameters equal to the mean of this distribution ($\vm$) is similar to standard neural network training but $\vSigma$ also provides an estimate of its stability.
To be precise, for each parameter $m_i$ the variance $\Sigma_{ii}$ indicates how much this parameter can be changed without significant performance degradation \revision{which can be seen as a measure of uncertainty}.
{We also use multiple models obtained from \revision{independent} IVON training runs to form a Deep Ensemble~\citep{lakshminarayanan2017simple} in order to study multimodal \revision{token- or sequence-level} posteriors. This can be seen as constructing a mixture-of-Gaussian posterior with equal mixture component weights \revision{but incurs training overhead, since training time increases linearly with the number of mixture components.}}
\revision{Unless otherwise stated, we use four models in total for MBR, i.e. $|\mathcal{M}|=4$. For deep ensembles, we use the mean of each training run and for the unimodal method using IVON we use four samples from the posterior.}
\revision{For smaller models we train all parameters but for larger models we only train newly-inserted LoRA parameters $\vparam^\prime \in \real^{e}$, following IVON-LoRA~\citet{cong2024variational}. IVON-LoRA then learns a distribution $q(\vparam^\prime) \coloneqq \gauss(\vparam^\prime\mid\vm^\prime, \vSigma^\prime)$ while the original pretrained model parameters $\vparam$ remain fixed.}

\begin{table*}[t!]
    \centering
    \setlength{\tabcolsep}{3pt}
\def\arraystretch{1.05}
    \resizebox{\linewidth}{!}{\begin{tabular}{lccccccccccccccc}
         
        &\multicolumn{3}{c}{ IWSLT17 En-De}&\multicolumn{3}{c}{ WMT18 Tr-En} & \multicolumn{3}{c}{XSUM} & \multicolumn{2}{c}{ SAMSum} & \multicolumn{2}{c}{E2E NLG} & \multicolumn{1}{c}{STS-B}  \\
         Method & BLEU & COMET & LaBSE & BLEU & COMET & LaBSE& R-1 & R-L & FactCC & R-1 & R-L & R-1 & R-L & RMSE \\
         \hline
         \; \revision{MBR (AdamW)} & \revision{19.93} & \revision{76.62} & \revision{83.47} & \revision{14.75} & \revision{78.20} & \revision{76.02} & \revision{\bf 33.63} & \revision{25.67} & \revision{27.50} & \revision{46.47} & \revision{36.21} & \revision{67.88} & \revision{44.41} & \revision{0.330} \\ %
         \; MBR@Mean & 19.73 & 76.60 & 83.51 &15.27 & 78.44 & 77.12 & 33.04 & 25.19 & 23.56 & 46.17 & 35.98 & 68.74 & 45.16 & 0.284\\
         \multicolumn{11}{l}{\rule{0pt}{0.1in}\bf Sequence-level - \cref{eq:hyp_set_concat_estimator}} \\
         \; Unimodal & 20.89 & 77.42 & 84.01 & 15.66 & 79.01 & {\bf 77.79} & 33.39 & {\bf 25.73} & 26.07 & 46.40 & 36.51  & 69.36 & 45.57 & 0.271 \\
         \; Deep Ensemble & {\bf 21.24} & {\bf 77.94} & {\bf 84.20} & 15.63 & 79.01 & 77.60 & 33.37 & 25.68 & 27.40 & {\bf 46.71} & {\bf 36.87} & {\bf 69.56} & {\bf 45.77} & {\bf 0.269} \\
         \multicolumn{11}{l}{\rule{0pt}{0.1in}\bf Sequence-level - \cref{eq:seq_level_mbr_estimator}} \\
         \; Unimodal & 21.08 & 77.63 & 83.96 & 15.46 & 78.84 & 77.35 & 33.05 & 25.46 & 27.50 & 46.21 & 36.44 & 69.13 & 45.38 & 0.271 \\
         \; Deep Ensemble & 21.20 & 77.91 & 84.04 & {\bf 15.69} & {\bf 79.10} & 77.56 & 33.10 & 25.50 & {\bf 32.86} & 46.14 &  36.48 &  69.19 & 45.31 & {\bf 0.269} \\
         \hline
    \end{tabular}}
    \caption{Sequence-level model combination to account for weight-uncertainty can improve the performance of a finetuned Gemma-2B model on various language generation and scoring tasks. Even simple posteriors that do not incur overhead during finetuning can give ``for-free'' improvements (unimodal). The number of total MBR comparisons is the same for all methods and each dataset. 
    MBR@mean denotes decoding with a single model that is the mean of a variational distribution.
    }
    \label{tab:llm}
\end{table*}
\subsection{Weight Uncertainty \& Decoding}
\begin{table*}[t!]
    \centering
    \resizebox{\linewidth}{!}{\begin{tabular}{lcccccccccc}
        &\multicolumn{4}{c}{ WMT14 En-De} & \multicolumn{4}{c}{ IWSLT14 De-En} & & \\
        & \multicolumn{2}{c}{Sampling} & \multicolumn{2}{c}{Beam Search} & \multicolumn{2}{c}{Sampling} & \multicolumn{2}{c}{Beam Search} & MBR & Effective\\
         Method & BLEU & COMET & BLEU & COMET & BLEU & COMET & BLEU & COMET & comparisons & beam size \\
         \hline
        \; MBR@Mean & 23.37 & 71.04 & 27.56 & 75.23 & 33.69 & 74.71 & 35.90 & 76.65 & 400 & 20\\
         & 24.30 & 72.15 & 27.53 & 75.18 & 34.53 & 75.18 & 36.07 & 76.76 & 1600 & 40\\
         \multicolumn{11}{l}{\rule{0pt}{0.1in}\bf Sequence-level - \cref{eq:hyp_set_concat_estimator}} \\
         \; Unimodal & 24.31 & 72.09 & 27.52 & 75.16 & 34.59 & 75.15 & 35.78 & 76.55 & 1600 & 40 \\
         \; Deep Ensemble & {\bf 24.70} & 72.39 & {\bf 28.99} & 76.02 & {\bf 36.03} & 75.79 & 38.30 & 78.01 & 1600 & 40 \\
         \multicolumn{11}{l}{\rule{0pt}{0.1in}\bf Sequence-level - \cref{eq:seq_level_mbr_estimator}} \\        
         \; Unimodal & 24.21 & 72.15 & 27.56 & 75.21 & 34.65 & 75.20 & 35.99 & 76.67 & 1600 & 80\\
        \; Deep Ensemble & 24.67 & {\bf 72.58} & 28.29 & 75.70 & {35.42} & {\bf 75.84} & 37.42 & 77.69 & 1600 & 80\\
        \multicolumn{11}{l}{\rule{0pt}{0.1in}\bf Token-level} \\
        \; Unimodal & 23.44 & 71.36 & 27.75 & 75.19 & 33.62 & 74.68 & 35.94 & 76.66 & 400 & 80\\
        \; Deep Ensemble & 23.95 & 71.58 & 28.98 & {\bf 76.08} & 34.61 & 75.06 & {\bf 38.56} & {\bf 78.31} & 400 & 80\\
       \hline
    \end{tabular}}
    \caption{
    Weight uncertainty improves decoding when training from scratch and using ancestral sampling and beam search. 
    More complex posteriors (Deep Ensemble) provide better improvements.
    We use Transformer$_\text{big}$ on WMT14 and Transformer$_\text{base}$ on IWSLT17. Effective beam size $=$ number of beams per model times number of models (we use four).
    \label{tab:main_results}
    }
\end{table*}
\begin{table}[t]
    \centering
    \setlength{\tabcolsep}{5pt}
\def\arraystretch{1.05}
    \resizebox{\textwidth}{!}{\begin{tabular}{lcccccccccc}
         & \multicolumn{4}{c}{IWSLT17 De-En}  & \multicolumn{2}{c}{WMT19 Cs-En} & \multicolumn{2}{c}{XSUM} &MBR & Effective \\
         & \multicolumn{2}{c}{2 Models} & \multicolumn{2}{c}{3 Models} & \multicolumn{2}{c}{2 Models} & \multicolumn{2}{c}{3 Models} & Comparisons & Beam size\\
         Method & BLEU & COMET & BLEU & COMET & BLEU & COMET & R-1 & R-L\\
         \hline
         Single Model & 24.59 & 80.24 & 24.59 & 80.24 & 28.65 & 82.95 & 26.99 & 19.05 & 100 & 10 \\
         Sequence-level - \cref{eq:hyp_set_concat_estimator} & {\bf 26.66} & {\bf 81.60} & {\bf 29.12} & {\bf 83.06} & {\bf 30.60} & {\bf 84.12} & {\bf 28.27} & {\bf 20.22} & 400/900 & 20/30 \\
         Sequence-level - \cref{eq:seq_level_mbr_estimator} & 26.02 & 81.47 & 26.50 & 81.86 & 30.25 & 83.99 & 27.43 & 19.33 & 200/300 & 20/30 \\ 
         \hline
    \end{tabular}}
    \caption{Sequence-level model combination is also useful for ensembling zero-shot prompted LLMs. \cref{eq:hyp_set_concat_estimator} performs better but requires more computation.
    }
    \label{tab:llm_ensembling}
\end{table}
\label{sec:experiments_main}

\paragraph{Weight uncertainty improves decoding.}
{
\cref{tab:llm} and \cref{tab:main_results} show results using finetuned Gemma-2B and Transformer models that were pretrained from scratch, respectively, on various language generation and scoring benchmarks.
Results on two low-resource tasks from afroMT are found in~\cref{app:afromt}.
For a fair comparison, we match the number of MBR comparisons, i.e. evaluations of the utility function $u$ for the estimator, with the single-model MBR baseline, as described in~\cref{app:experimental_details_hyp_sizes}.

We find {\revision{in \cref{tab:llm} and \cref{tab:main_results}}} that weight uncertainty improves performance across all benchmarks, even with matched compute budgets.
\revision{That is, using just one model (and thereby neglecting weight uncertainty) performs worse than using multiple models sampled from the posterior and averaging their predictions. Improvements also tend to hold when compared to training with AdamW (\cref{tab:llm})}.
In particular, when using~\cref{eq:hyp_set_concat_estimator} with unimodal posteriors both training time and time needed for decoding are the same as for the single-model MBR baseline.
\revision{We ensure that the time needed for decoding is the same by i) using only as many MBR comparisons as MBR@mean for our methods and ii) always using the same or smaller effective beam size, which is measured by the number of beams per model multiplied the number of models. We validate this empirically in~\cref{app:decoding_time}.}
Not only do results improve when using word-overlap metrics like BLEU, but also when using quality estimation (COMET) and hallucination metrics (LaBSE).
Notably, on IWSLT17 all improvements observed in COMET score when using uncertainty-aware vs. standard MBR indicate there is an {\revision{estimated}} $>$85\% chance that humans would distinguish the former system as better---as per~\citet{kocmi-etal-2024-navigating}.
Improvements also hold for the STS-B sentence similarity scoring task.
The estimators of~\cref{eq:hyp_set_concat_estimator} and~\cref{eq:seq_level_mbr_estimator} perform similarly even though~\cref{eq:hyp_set_concat_estimator} uses a smaller hypothesis set size than~\cref{eq:seq_level_mbr_estimator}.
}

\paragraph{Comparison of uni- and multimodal posteriors.}
{Next, we compare unimodal posteriors that can be learned without overhead during training to multimodal posteriors based on Deep Ensembles.
Such posteriors incur significant overhead during training, because one separate training run with different initialization and data order is required per ensemble member, but can incorporate knowledge from different loss basins---a characteristic that has proven to be beneficial~\citep{lion2023how}.

When training from scratch (\cref{tab:main_results}),  unimodal posteriors do not consistently outperform the single model baseline when compute budgets are equivalent. In contrast, multimodal Deep Ensemble posteriors can deliver significant improvements. 
On the other hand, when finetuning (\cref{tab:llm}), unimodal posteriors can provide strong improvements, performing on par with Deep Ensembles.
We hypothesize that this difference can be attributed to the use of LoRA for finetuning---which explores a smaller subspace of potential posterior parameters and may therefore pose a comparably easier learning problem than estimating the variance of a posterior over all parameters. %
Further, finetuning may not work that well for Deep Ensembles due to the models still landing in the same basin~\citep{pmlr-v119-frankle20a}.
We connect our findings to prediction diversity in~\cref{sec:quality_diversity}.}

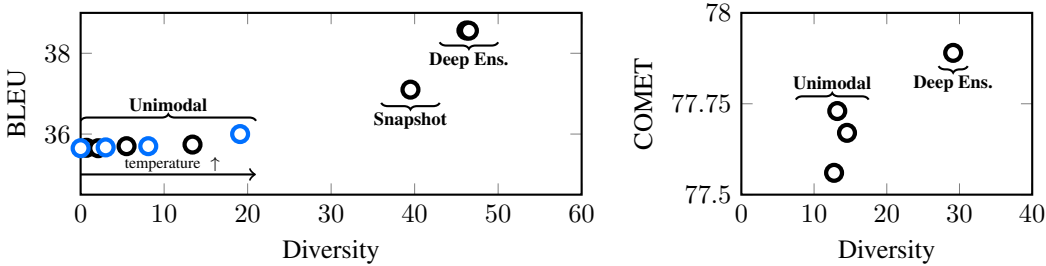
\begin{figure}[t!]
\centering
\begin{subfigure}{.59\textwidth}
\begin{tikzpicture}
\begin{axis}[
    xlabel=Diversity,
    ylabel=BLEU,
    xmin=0, xmax=60,
    ymin=34.5, ymax=39,
    height=4cm,
    width=\linewidth,
    xtick={0.0,10, ..., 100},
    ytick={30,32,...,40},
    style=thick,
    xlabel near ticks,
    ylabel near ticks,
    legend style={at={(0.025,0.975)}, anchor=north west}
            ]
\addplot+[ultra thick,only marks,mark=*,black, mark size=3pt, mark options={scale=1, fill=white}] plot coordinates {
(0.7, 35.66)
(2.1, 35.65)
(5.5, 35.70)
(13.4, 35.74)
(39.5, 37.1)
(46.2, 38.564)
(46.5, 38.56)
};
\addplot+[ultra thick,only marks,mark=*,brandeisblue, mark size=3pt, mark options={scale=1, fill=white}] plot coordinates {
(0.0, 35.65)
(3.0, 35.67)
(8.1, 35.70)
(19.1, 36.00)
};
\draw [thick,decoration={brace,raise=5pt},decorate] 
  (axis cs:0,36) --
    node[above=5pt] {\scriptsize \bf Unimodal} 
  (axis cs:21,36);
\draw [thick,
decoration={brace,mirror,raise=5pt},decorate
] 
  (axis cs:36,37.1) --
    node[below=5pt] {\scriptsize \bf Snapshot} 
  (axis cs:43,37.1);
  \draw [thick,decoration={brace,mirror, raise=5pt},decorate] 
  (axis cs:43,38.6) --
    node[below=5pt] {\scriptsize \bf Deep Ens.} 
  (axis cs:50,38.6);
\draw [->] 
  (axis cs:0,35) --
    node[above=-2.5pt] {\tiny \; $\text{temperature } \uparrow$} 
  (axis cs:21,35);
\end{axis}
\end{tikzpicture}
\end{subfigure}
\begin{subfigure}{.39\textwidth}
\begin{tikzpicture}
\begin{axis}[
    xlabel=Diversity,
    ylabel=COMET,
    xmin=0, xmax=40,
    ymin=77.5, ymax=78.0,
    height=4cm,
    width=\linewidth,
    xtick={0.0,10, ..., 100},
    ytick={77.5,77.75,78.0,...,100},
    style=thick,
    xlabel near ticks,
    ylabel near ticks,
    legend style={at={(0.025,0.975)}, anchor=north west}
            ]
\addplot+[ultra thick,only marks,mark=*,black, mark size=3pt, mark options={scale=1, fill=white}] plot coordinates {
(13.19,77.73)
(12.75,77.56)
(14.53, 77.67)
(29.13,77.89)
};
\addplot+[ultra thick,only marks,mark=*,brandeisblue, mark size=3pt, mark options={scale=1, fill=white}] plot coordinates {
};
\draw [thick,decoration={brace,raise=5pt},decorate] 
  (axis cs:7.5,77.725) --
    node[above=5pt] {\scriptsize \bf Unimodal} 
  (axis cs:17.5,77.725);
  \draw [thick,decoration={brace,mirror, raise=5pt},decorate] 
  (axis cs:27.13,77.89) --
    node[below=5pt] {\scriptsize \bf Deep Ens.} 
  (axis cs:31.13,77.89);
\end{axis}
\end{tikzpicture}
\end{subfigure}
\caption{Our methods are more successful when the ensembled models are diverse. We compare a unimodal to mixture-based posteriors using Snapshot Ensembles and Deep Ensembles.
Sampling from a unimodal posterior with higher temperature can increase diversity and improve performance ({\color{brandeisblue}in blue}).
Left: token-level combination on IWSLT14 using beam search and Transformer$_\text{base}$. Right: sequence-level combination (\cref{eq:seq_level_mbr_estimator}) on IWSLT17 using ancestral sampling and Gemma-2B.
}
\label{fig:quality_diversity}
\end{figure}

\paragraph{Comparison of sequence- and token-level posteriors.}
Here, we compare the use of sequence- and token-level posteriors (\cref{eq:tok_level_mbr,eq:hyp_set_concat_estimator,eq:seq_level_mbr_estimator}) in MBR. 
\cref{tab:main_results} shows that improvements over the baseline with token-level combination are much stronger when using beam search instead of ancestral sampling to create hypothesis sets\footnote{
Beam search provides a biased estimate and is similar to sampling from a low-temperature distribution.}.
When using a mixture-based posterior, performance is improved in both settings.
Sequence-level combination, on the other hand, provides similar improvements for both settings, with~\cref{eq:hyp_set_concat_estimator} providing similar results to token-level aggregation.
Hence, the preferred method may also depend on the decoding algorithm used to create the hypothesis set. 

\paragraph{Ensembling zero-shot models.}
\cref{tab:llm_ensembling} shows results obtained when ensembling the outputs of various zero-shot prompted LLMs on IWSLT17 De-En with a hypothesis set size of 10.
We compare the estimator using an additive union of hypothesis sets (\cref{eq:hyp_set_concat_estimator}) to using a soft model average (\cref{eq:seq_level_mbr_estimator}) and the average single model performance.
Both estimators are effective for ensembling but \cref{eq:hyp_set_concat_estimator} performs best, albeit with the highest computational complexity. Details are in~\cref{app:zero_shot}.

\subsection{Correlation of Quality and Diversity}
\label{sec:quality_diversity}
Next, we show that the performance of MBR with weight-uncertainty is correlated with the prediction diversity of ensembled models, potentially, due to incorporating knowledge from multiple loss basins.
This is in line with prior works on ensembling which have found that diversity is important for good performance~\citep{fort2019deep, masegosa2020learning} but can form a trade-off with individual model performance~\citep{abe2022the, wood2023unified}.

We empirically validate this in~\cref{fig:quality_diversity}, where we plot BLEU and COMET on IWSLT14 and IWSLT17 against the prediction diversity. We measure diversity as $100$ minus average self-BLEU; self-BLEU scores are measured on the set of greedy decoding outputs of each ensemble member, similar to~\citet{shen2019mixture}.
For finetuning, the models from the unimodal posterior are more diverse than when pretraining.
The plot shows a clear correlation between both metrics.
We ask two questions: 1) can diversity be promoted in unimodal pretrained posteriors to improve performance and 2) can we find a method with the same pretraining overhead as a unimodal posterior but more expressiveness?

For the first, note that the variance of the IVON posterior is $\vsigma^2 = 1/\lambda(\vh+\delta)$, where $\vh$ is the expected Hessian of the loss, $\delta$ is weight-decay and $\lambda$ the effective sample size which can be seen as an (inverse) temperature parameter.
We decrease $\lambda$ gradually, which samples models from the posterior with higher temperature.
This improves diversity and can improve performance.
For the latter, we use a mixture-of-Gaussian consisting of checkpoints from one training run, denoted by ``snapshot''~\citep{huang2017snapshot}.
This comes at no training time increase but can improve performance by incorporating knowledge from different regions along the optimization trajectory, \revision{as shown in~\cref{fig:quality_diversity}}.

\subsection{Selective Prediction with Bayes Risk}
\label{sec:selective_prediction}
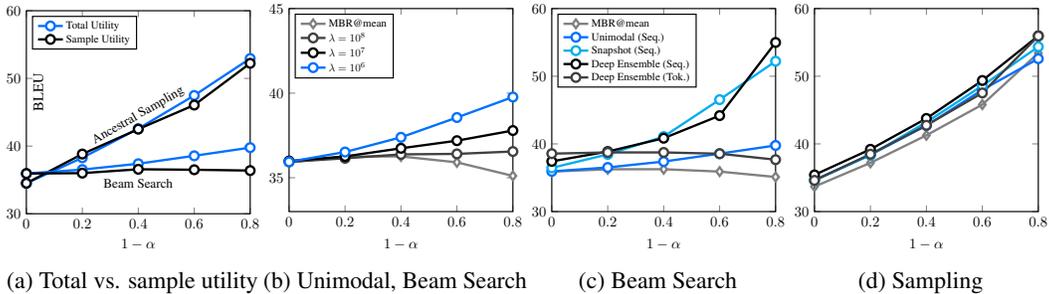
\begin{figure*}[t!]
\centering
\begin{subfigure}{.25\linewidth}
        \centering
\resizebox{\columnwidth}{!}{\begin{tikzpicture}
\begin{axis}[
    xlabel={$1-\alpha$},
    ylabel=BLEU,
    xmin=0.0, xmax=0.8,
    ymin=30, ymax=60.0,
    height=6.5cm,
    width=7cm,
    xtick={0.0,0.2, ..., 0.8},
    ytick={30,40,...,60},
    style=thick,
    xlabel near ticks,
    ylabel near ticks,
    y label style={at={(0.1,0.6)}},
    legend style={at={(0.025,0.975)}, anchor=north west, nodes={scale=0.75, transform shape}},
    legend cell align={left}
            ]
\addplot+[ultra thick,mark=*,brandeisblue, mark size=3pt, mark options={scale=1, fill=white}] plot coordinates {
(0.0, 34.51)
(0.2, 38.27)
(0.4, 42.57)
(0.6, 47.48)
(0.8, 52.94)
};
\node[rotate=30] at (axis cs:0.4,44.5) {\small Ancestral Sampling};
\addplot+[ultra thick,mark=*,black, mark size=3pt, mark options={scale=1, fill=white}] plot coordinates {
(0.0, 34.51)
(0.2, 38.82)
(0.4, 42.5)
(0.6, 46.06)
(0.8, 52.23)
};
\addplot+[ultra thick,mark=*, brandeisblue, mark size=3pt, mark options={scale=1, fill=white}] plot coordinates {
(0.0, 35.94)
(0.2, 36.52)
(0.4, 37.39)
(0.6, 38.56)
(0.8, 39.77)
};
\node[] at (axis cs:0.4,34.5) {\small Beam Search};
\addplot+[ultra thick,mark=*,black, mark size=3pt, mark options={scale=1, fill=white}] plot coordinates {
(0.0, 35.94)
(0.2, 36.0)
(0.4, 36.57)
(0.6, 36.5)
(0.8, 36.38)
};
\addlegendentry{Total Utility}
\addlegendentry{Sample Utility}
\end{axis}
\end{tikzpicture}}
\caption{Total vs. sample utility}
\end{subfigure}\begin{subfigure}{.25\linewidth}
        \centering
\resizebox{\columnwidth}{!}{\begin{tikzpicture}
\begin{axis}[
    xlabel={$1-\alpha$},
    xmin=0.0, xmax=0.8,
    ymin=33, ymax=45.0,
    height=6.5cm,
    width=7cm,
    xtick={0.0,0.2, ..., 0.8},
    ytick={35,40,...,45},
    style=thick,
    ylabel near ticks,
    legend style={at={(0.025,0.975)}, anchor=north west, nodes={scale=0.75, transform shape}},
    legend cell align={left}
            ]
\addplot+[ultra thick,mark=diamond*,gray, mark size=3pt, mark options={scale=1, fill=white}] plot coordinates {
(0.0, 35.91)
(0.2, 36.25)
(0.4, 36.26)
(0.6, 35.91)
(0.8, 35.11)
};
\addplot+[ultra thick,mark=*,darkgray, mark size=3pt, mark options={scale=1, fill=white}] plot coordinates {
(0.0, 35.94)
(0.2, 36.15) 
(0.4, 36.38)
(0.6, 36.41)
(0.8, 36.55)
};
\addplot+[ultra thick,mark=*,black, mark size=3pt, mark options={scale=1, fill=white}] plot coordinates {
(0.0, 35.99) 
(0.2, 36.26) 
(0.4, 36.74)
(0.6, 37.19)
(0.8, 37.79)
};
\addplot+[ultra thick,mark=*,brandeisblue, mark size=3pt, mark options={scale=1, fill=white}] plot coordinates {
(0.0, 35.94)
(0.2, 36.52)
(0.4, 37.39)
(0.6, 38.56)
(0.8, 39.77)
};
\addlegendentry{MBR@mean}
\addlegendentry{$\lambda = 10^8$}
\addlegendentry{$\lambda = 10^7$}
\addlegendentry{$\lambda = 10^6$}
\end{axis}
\end{tikzpicture}}
\caption{Unimodal, Beam Search}
\end{subfigure}\begin{subfigure}{.25\linewidth}
        \centering
\resizebox{\columnwidth}{!}{\begin{tikzpicture}
\begin{axis}[
    xlabel={$1-\alpha$},
    xmin=0.0, xmax=0.8,
    ymin=30, ymax=60.0,
    height=6.5cm,
    width=7cm,
    xtick={0.0,0.2, ..., 0.8},
    ytick={30,40,...,60},
    style=thick,
    ylabel near ticks,
    legend style={at={(0.025,0.975)}, anchor=north west, nodes={scale=0.75, transform shape}},
    legend cell align={left}
            ]
\addplot+[ultra thick,mark=diamond*,gray, mark size=3pt, mark options={scale=1, fill=white}] plot coordinates {
(0.0, 35.91)
(0.2, 36.25)
(0.4, 36.26)
(0.6, 35.91)
(0.8, 35.11)
};
\addplot+[ultra thick,mark=*,brandeisblue, mark size=3pt, mark options={scale=1, fill=white}] plot coordinates {
(0.0, 35.94)
(0.2, 36.52)
(0.4, 37.39)
(0.6, 38.56)
(0.8, 39.77)
};
\addplot+[ultra thick,mark=*,cyan, mark size=3pt, mark options={scale=1, fill=white}] plot coordinates {
(0.0, 36.46)
(0.2, 38.45) 
(0.4, 41.08)
(0.6, 46.54)
(0.8, 52.22)
};
\addplot+[ultra thick,mark=*,black, mark size=3pt, mark options={scale=1, fill=white}] plot coordinates {
(0.0, 37.42)
(0.2, 38.87) 
(0.4, 40.84)
(0.6, 44.18)
(0.8, 54.99)
};
\addplot+[ultra thick,mark=*,darkgray, mark size=3pt, mark options={scale=1, fill=white}] plot coordinates {
(0.0, 38.56) 
(0.2, 38.74) 
(0.4, 38.74)
(0.6, 38.55)
(0.8, 37.67)
};
\addlegendentry{MBR@mean}
\addlegendentry{Unimodal (Seq.)}
\addlegendentry{Snapshot (Seq.)}
\addlegendentry{Deep Ensemble (Seq.)}
\addlegendentry{Deep Ensemble (Tok.)}
\end{axis}
\end{tikzpicture}}
\caption{Beam Search}
\end{subfigure}\begin{subfigure}{.25\linewidth}
        \centering
\resizebox{\columnwidth}{!}{\begin{tikzpicture}
\begin{axis}[
    xlabel={$1-\alpha$},
    xmin=0.0, xmax=0.8,
    ymin=30, ymax=60.0,
    height=6.5cm,
    width=7cm,
    xtick={0.0,0.2, ..., 0.8},
    ytick={30,40,...,60},
    style=thick,
    ylabel near ticks,
    legend style={at={(0.025,0.975)}, anchor=north west}
            ]
\addplot+[ultra thick,mark=diamond,gray, mark size=3pt, mark options={scale=1, fill=white}] plot coordinates {
(0.0, 33.69)
(0.2, 37.2)
(0.4, 41.27)
(0.6, 45.81)
(0.8, 53.42)
};
\addplot+[ultra thick,mark=*,brandeisblue, mark size=3pt, mark options={scale=1, fill=white}] plot coordinates {
(0.0, 34.65) 
(0.2, 38.39)
(0.4, 42.71)
(0.6, 47.99)
(0.8, 52.59)
};
\addplot+[ultra thick,mark=*,cyan, mark size=3pt, mark options={scale=1, fill=white}] plot coordinates {
(0.0, 34.61) 
(0.2, 38.35)
(0.4, 43.21)
(0.6, 48.6)
(0.8, 54.36)
};
\addplot+[ultra thick,mark=*,black, mark size=3pt, mark options={scale=1, fill=white}] plot coordinates {
(0.0, 35.42) 
(0.2, 39.2)
(0.4, 43.76)
(0.6, 49.37)
(0.8, 55.96)
};
\addplot+[ultra thick,mark=*,darkgray, mark size=3pt, mark options={scale=1, fill=white}] plot coordinates {
(0.0, 34.61) 
(0.2, 38.48) 
(0.4, 42.75)
(0.6, 47.53)
(0.8, 55.94)
};
\end{axis}
\end{tikzpicture}}
\caption{Sampling}
\end{subfigure}
    \caption{
    Total risk and best-output-risk are useful for selective prediction. (a) Creating hypothesis sets with sampling performs better than beam search. (b) Increasing temperature when sampling from unimodal posteriors improves selective prediction. (c) When using beam search more Deep Ensembles work best. (d) For sampling, all methods work well. Results on IWSLT14 with Transformer$_\text{base}$.}
    \label{fig:selective_prediction}
\end{figure*}

Here, we explore the use of expected Bayes risk for selective prediction on IWSLT14. We observe that both the maximum output utility and the expected output utility (i.e., average expected utility across outputs) can be used effectively for selective prediction.
Our results are summarized in~\cref{fig:selective_prediction}.

First, we find in~\cref{fig:selective_prediction} (a) that using the \revision{total} expected utility for selective prediction performs slightly better than \revision{just using the expected utility of the chosen output.} This is especially true when creating hypothesis sets with beam search, which performs much worse than ancestral sampling.
Next, we again sample from the unimodal posterior with different temperatures (via decreasing $\lambda$).
We find that this improves selective prediction with MBR when using beam search (\cref{fig:selective_prediction} (b)).

Finally, we evaluate the influence of the posterior approximation.
First, we find that a hypothesis set built with ancestral sampling is reliable independent of the used posterior.
Even the single model baseline works well but is outperformed by using an ensemble and more expressive posteriors give bigger improvements.
For beam search, the baseline completely fails and token-level combination can be unreliable.
Sequence-level combination (\cref{eq:seq_level_mbr_estimator}) performs much better, especially with more expressive multimodal posteriors.
These results are shown in \cref{fig:selective_prediction} (c, d).

\subsection{Scaling Behavior}
\label{sec:scaling}
\begin{figure*}[t!]
\centering
\begin{subfigure}{.25\linewidth}
        \centering
\resizebox{\columnwidth}{!}{\begin{tikzpicture}
\begin{axis}[
    xlabel=Ensemble size,
    ylabel=BLEU,
    xmin=0.0, xmax=33.0,
    ymin=33, ymax=37,
    height=7cm,
    height=6.5cm,
    xtick={2,4,8,16,32},
    xmode=log,log basis x=2,
    ytick={30,32,...,60},
    style=thick,
    xlabel near ticks,
    ylabel near ticks,
    y label style={at={(0.1,0.75)}},
    legend style={at={(0.025,0.975)}, anchor=north west, nodes={scale=0.75, transform shape}},
    legend cell align={left}
            ]
\addplot+[ultra thick,mark=square*,brandeisblue, mark size=3pt, mark options={scale=1, fill=white}] plot coordinates {
(2, 33.51)
(4, 33.62)
(8, 33.62)
(16, 33.68)
(32, 33.88)
};
\addplot+[ultra thick,mark=*,brandeisblue, mark size=3pt, mark options={scale=1, fill=white}] plot coordinates {
(2, 34.06)
(4, 34.61)
(8, 35.03)
(16, 35.17)
(32, 34.94)
};

\addplot+[ultra thick,mark=square*,black, mark size=3pt, mark options={scale=1, fill=white}] plot coordinates {
(2, 34.02)
(4, 34.65)
(8, 34.59)
(16, 34.56)
(32, 34.5)
};
\addplot+[ultra thick,mark=*,black, mark size=3pt, mark options={scale=1, fill=white}] plot coordinates {
(2, 34.44)
(4, 35.42)
(8, 35.84)
(16, 36.16)
(32, 36.54)
};
\node[rotate=17.5] at (axis cs:12.1,36.3) {\small Sequence-level};
\node[rotate=9] at (axis cs:12,35.335) {\small \color{brandeisblue} Token-level};

\end{axis}
\end{tikzpicture}}
\caption{Ancestral Sampling}
\end{subfigure}\begin{subfigure}{.25\linewidth}
         \centering
\resizebox{\columnwidth}{!}{\begin{tikzpicture}
\begin{axis}[
    xlabel=Ensemble size,
    xmin=0.0, xmax=33.0,
    ymin=35, ymax=40.0,
    height=7cm,
    height=6.5cm,
    xtick={2,4,8,16,32},
    xmode=log,log basis x=2,
    ytick={30,32,...,60},
    style=thick,
    xlabel near ticks,
    ylabel near ticks,
    y label style={at={(0.1,0.75)}},
    legend style={at={(0.025,0.975)}, anchor=north west, nodes={scale=0.75, transform shape}},
    legend cell align={left}
            ]
\addplot+[ultra thick,mark=square*,brandeisblue, mark size=3pt, mark options={scale=1, fill=white}] plot coordinates {
(2, 35.92)
(4, 35.94)
(8, 35.93)
(16, 35.92)
(32, 35.92)
};
\addplot+[ultra thick,mark=*,brandeisblue, mark size=3pt, mark options={scale=1, fill=white}] plot coordinates {
(2, 37.65)
(4, 38.56)
(8, 38.84)
(16, 39.05)
(32, 39.02)
};
\addplot+[ultra thick,mark=square*,black, mark size=3pt, dashed, mark options={scale=1, fill=white}] plot coordinates {
(2, 35.91)
(4, 35.94)
(8, 35.92)
(16, 35.92)
(32, 35.91)
};
\addplot+[ultra thick,mark=*,black, mark size=3pt, mark options={scale=1, fill=white}] plot coordinates {
(2, 36.75)
(4, 37.08)
(8, 37.62)
(16, 37.93)
(32, 38.0)
};
\end{axis}
\end{tikzpicture}}
\caption{Beam Search}
\end{subfigure}\begin{subfigure}{.25\linewidth}
           \centering
\resizebox{\columnwidth}{!}{\begin{tikzpicture}
\begin{axis}[
    xlabel=Beam size,
    xmin=10.0, xmax=70.0,
    ymin=32, ymax=38.0,
    height=7cm,
    height=6.5cm,
    xtick={0, 10,..., 70},
    ytick={30,32,...,60},
    style=thick,
    xlabel near ticks,
    ylabel near ticks,
    y label style={at={(0.1,0.75)}},
    legend style={at={(0.025,0.975)}, anchor=north west, nodes={scale=0.75, transform shape}},
    legend cell align={left}
            ]
\addplot+[ultra thick,mark=*,gray, mark size=3pt, mark options={scale=1, fill=white}] plot coordinates {
(10, 32.55)
(20, 33.59)
(30, 34.13)
(40, 34.53)
(50, 34.7)
(60, 34.88)
(70, 35.03)
};
\addplot+[ultra thick,mark=square*,brandeisblue, mark size=3pt,  mark options={scale=1, fill=white}] plot coordinates {
(10, 32.48)
(20, 33.62)
(30, 34.13)
(40, 34.58)
(50, 34.63)
(60, 34.96)
(70, 35.0)
};
\addplot+[ultra thick,mark=*,brandeisblue, mark size=3pt, mark options={scale=1, fill=white}] plot coordinates {
(10, 33.1)
(20, 34.61)
(30, 35.45)
(40, 35.83)
(50, 36.18)
(60, 36.36)
(70, 36.55)
};
\addplot+[ultra thick,mark=square*,black, mark size=3pt, mark options={scale=1, fill=white}] plot coordinates {
(10, 33.69)
(20, 34.65)
(30, 34.7)
(40, 35.05)
(50, 35.02)
(60, 35.13)
(70, 35.11)
};
\addplot+[ultra thick,mark=*,black, mark size=3pt, mark options={scale=1, fill=white}] plot coordinates {
(10, 34.47)
(20, 35.42)
(30, 35.73)
(40, 35.71)
(50, 35.95)
(60, 36.03)
(70, 36.13)
};
\end{axis}
\end{tikzpicture}}
\caption{Ancestral Sampling}
\end{subfigure}\begin{subfigure}{.25\linewidth}
           \centering
\resizebox{\columnwidth}{!}{\begin{tikzpicture}
\begin{axis}[
    xlabel=Beam size,
    xmin=10.0, xmax=70.0,
    ymin=35, ymax=40.0,
    height=7cm,
    height=6.5cm,
    xtick={0, 10,..., 70},
    ytick={30,32,...,60},
    style=thick,
    xlabel near ticks,
    ylabel near ticks,
    y label style={at={(0.1,0.75)}},
    legend style={at={(0.025,0.975)}, anchor=north west, nodes={scale=0.75, transform shape}},
    legend cell align={left}
            ]
\addplot+[ultra thick,mark=*,gray, mark size=3pt, mark options={scale=1, fill=white}] plot coordinates {
(10, 35.77)
(20, 35.91)
(30, 35.99)
(40, 36.07)
(50, 36.14)
(60, 36.12)
(70, 36.17)
};
\addplot+[ultra thick,mark=square*,brandeisblue, mark size=3pt,  mark options={scale=1, fill=white}] plot coordinates {
(10, 35.78)
(20, 35.94)
(30, 35.97)
(40, 36.1)
(50, 36.15)
(60, 36.11)
(70, 36.14)
};
\addplot+[ultra thick,mark=*,brandeisblue, mark size=3pt, mark options={scale=1, fill=white}] plot coordinates {
(10, 38.43)
(20, 38.56)
(30, 38.61)
(40, 38.63)
(50, 38.59)
(60, 38.65)
(70, 38.65)
};
\addplot+[ultra thick,mark=square*,black, mark size=3pt, mark options={scale=1, fill=white}] plot coordinates {
(10, 35.91)
(20, 35.99)
(30, 35.98)
(40, 36.07)
(50, 36.10)
(60, 36.09)
(70, 36.15)
};
\addplot+[ultra thick,mark=*,black, mark size=3pt, mark options={scale=1, fill=white}] plot coordinates {
(10, 37.17)
(20, 37.42)
(30, 37.48)
(40, 37.62)
(50, 37.67)
(60, 37.7)
(70, 37.77)
};
\end{axis}
\end{tikzpicture}}
\caption{Beam Search}
\end{subfigure}
    \caption{
    Scaling behavior on IWSLT14 with Transformer$_\text{base}$ in terms of ensemble (a, b) and hypothesis set size (c, d).
    (a, b) For a unimodal posterior ($\square$), larger ensembles improve token-level combination using sampling but not beam search. For Deep Ensemble posteriors ($\circ$), larger ensembles generally improve performance.
    (c, d) Sequence-level combination (\cref{eq:seq_level_mbr_estimator}) performs better for smaller beam sizes but is outperformed by token-level combination at larger ones.
    Scaling the hypothesis set produces stronger improvements for ancestral sampling than beam search.
    \label{fig:scaling}
    }
\end{figure*}
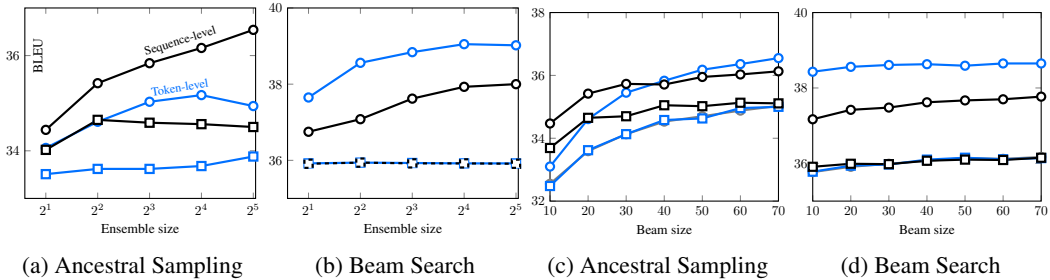

Lastly, we examine the scaling behavior of token- and sequence-level combination (\cref{eq:seq_level_mbr_estimator}) with different posteriors.
Results are summarized in~\cref{fig:scaling}.
First, we show scaling the ensemble size in~\cref{fig:scaling} (a) for ancestral sampling and beam search (b).
Using beam search, both token- (in blue) and sequence-level (in black) combination using unimodal posteriors provide no improvements.
For ancestral sampling, we find improvements with a unimodal posterior, especially at larger ensemble sizes of $32$ models, but sequence-level combination of a unimodal posterior only improves until 4 models.
In all other settings, scaling the ensemble size is usually beneficial.

When scaling hypothesis sets with beam search, the improvements are small, likely because the hypothesis sets lack diversity.
Ancestral sampling shows a different picture and we obtain strong improvements when scaling hypothesis sets.
For small hypothesis sets it is better to use sequence-level ensembling but for larger sizes token-level combination can be better.

\section{Conclusion}
In this work, we explore using a Minimum Bayes Risk approach to account for weight uncertainty in language model decoding.
We investigate different methods \revision{which} combine predictions from multiple models \revision{either} during generation or afterwards. \revision{Importantly, the latter can be used to ensemble any set of LLMs}.
We benchmark the methods on different language generation and scoring tasks for prompted \revision{and} finetuned models, \revision{as well as models trained from scratch. We} show that weight uncertainty can effectively improve decoding.
We evaluate the effects of using different posterior distributions. More complex distributions can sometimes provide stronger performance improvements but also simple methods can improve performance. \revision{Crucially, the improvements with simpler methods do not incur training or test-time overhead.}
\revision{We also connect our findings to} prediction diversity, \revision{which} is important for both standard MBR and when using its expected utility for selective prediction, \revision{and show that improvements scale with model and sample sizes.}
Overall, we find that the uncertainty-aware variant of MBR proposed in this paper leads to better and more robust language generation. 
\revision{Altogether, our method provides a principled approach for scaling test-time compute.}

\section*{Ethics Statement}
Our work uses probabilistic language models to generate language. 
Even when used with care, such models can produce outputs that are, among others, harmful, toxic, and hallucinated and our methods can not guarantee that such outputs are not generated.
However, we aim to improve the robustness of language generation methods and, therefore, aim to alleviate these issues.
Therefore, we believe there to be no direct ethical concern in our work.

\section*{\revision{Acknowledgements}}
\revision{This project has received funding by the German Federal Ministry of Education and Research and the Hessian Ministry of Higher Education, Research, Science and the Arts within their joint support of the National Research Center for Applied Cybersecurity ATHENE. 
Clara Meister was supported by a Google PhD Fellowship. 
This work is supported by JST CREST Grant Number JP-MJCR2112.

We thank Seyed Arshan Dalili for help with running the LLM experiments in~\cref{tab:llm_ensembling}.
}

\bibliography{anthology, iclr2024_conference}

\begin{thebibliography}{89}
\providecommand{\natexlab}[1]{#1}
\providecommand{\url}[1]{\texttt{#1}}
\expandafter\ifx\csname urlstyle\endcsname\relax
  \providecommand{\doi}[1]{doi: #1}\else
  \providecommand{\doi}{doi: \begingroup \urlstyle{rm}\Url}\fi

\bibitem[Abe et~al.(2022)Abe, Buchanan, Pleiss, and Cunningham]{abe2022the}
Taiga Abe, E.~Kelly Buchanan, Geoff Pleiss, and John~Patrick Cunningham.
\newblock The best deep ensembles sacrifice predictive diversity.
\newblock In \emph{I Can't Believe It's Not Better Workshop: Understanding Deep Learning Through Empirical Falsification}, 2022.
\newblock URL \url{https://openreview.net/forum?id=6sBiAIpkUiO}.

\bibitem[Alquier(2024)]{alquier2021user}
Pierre Alquier.
\newblock User-friendly introduction to pac-bayes bounds.
\newblock \emph{Foundations and Trends® in Machine Learning}, 17\penalty0 (2):\penalty0 174--303, 2024.
\newblock ISSN 1935-8237.
\newblock \doi{10.1561/2200000100}.
\newblock URL \url{http://dx.doi.org/10.1561/2200000100}.

\bibitem[Barrault et~al.(2019)Barrault, Bojar, Costa-juss{\`a}, Federmann, Fishel, Graham, Haddow, Huck, Koehn, Malmasi, Monz, M{\"u}ller, Pal, Post, and Zampieri]{barrault-etal-2019-findings}
Lo{\"\i}c Barrault, Ond{\v{r}}ej Bojar, Marta~R. Costa-juss{\`a}, Christian Federmann, Mark Fishel, Yvette Graham, Barry Haddow, Matthias Huck, Philipp Koehn, Shervin Malmasi, Christof Monz, Mathias M{\"u}ller, Santanu Pal, Matt Post, and Marcos Zampieri.
\newblock Findings of the 2019 conference on machine translation ({WMT}19).
\newblock In Ond{\v{r}}ej Bojar, Rajen Chatterjee, Christian Federmann, Mark Fishel, Yvette Graham, Barry Haddow, Matthias Huck, Antonio~Jimeno Yepes, Philipp Koehn, Andr{\'e} Martins, Christof Monz, Matteo Negri, Aur{\'e}lie N{\'e}v{\'e}ol, Mariana Neves, Matt Post, Marco Turchi, and Karin Verspoor (eds.), \emph{Proceedings of the Fourth Conference on Machine Translation (Volume 2: Shared Task Papers, Day 1)}, pp.\  1--61, Florence, Italy, August 2019. Association for Computational Linguistics.
\newblock \doi{10.18653/v1/W19-5301}.
\newblock URL \url{https://aclanthology.org/W19-5301}.

\bibitem[Bhandari \& Brennan(2023)Bhandari and Brennan]{bhandari-brennan-2023-trustworthiness}
Prabin Bhandari and Hannah Brennan.
\newblock Trustworthiness of children stories generated by large language models.
\newblock In C.~Maria Keet, Hung-Yi Lee, and Sina Zarrie{\ss} (eds.), \emph{Proceedings of the 16th International Natural Language Generation Conference}, pp.\  352--361, Prague, Czechia, September 2023. Association for Computational Linguistics.
\newblock \doi{10.18653/v1/2023.inlg-main.24}.
\newblock URL \url{https://aclanthology.org/2023.inlg-main.24}.

\bibitem[Blundell et~al.(2015)Blundell, Cornebise, Kavukcuoglu, and Wierstra]{blundell2015weight}
Charles Blundell, Julien Cornebise, Koray Kavukcuoglu, and Daan Wierstra.
\newblock Weight uncertainty in neural network.
\newblock In \emph{International conference on machine learning}, pp.\  1613--1622. PMLR, 2015.
\newblock URL \url{https://proceedings.mlr.press/v37/blundell15.html}.

\bibitem[Bojar et~al.(2014)Bojar, Buck, Federmann, Haddow, Koehn, Leveling, Monz, Pecina, Post, Saint-Amand, Soricut, Specia, and Tamchyna]{bojar14wmt}
Ondrej Bojar, Christian Buck, Christian Federmann, Barry Haddow, Philipp Koehn, Johannes Leveling, Christof Monz, Pavel Pecina, Matt Post, Herve Saint-Amand, Radu Soricut, Lucia Specia, and Ale~{s} Tamchyna.
\newblock Findings of the 2014 workshop on statistical machine translation.
\newblock In \emph{Proceedings of the Ninth Workshop on Statistical Machine Translation}, pp.\  12--58, Baltimore, Maryland, USA, June 2014. Association for Computational Linguistics.
\newblock URL \url{http://www.aclweb.org/anthology/W/W14/W14-3302}.

\bibitem[Bojar et~al.(2017)Bojar, Chatterjee, Federmann, Graham, Haddow, Huang, Huck, Koehn, Liu, Logacheva, Monz, Negri, Post, Rubino, Specia, and Turchi]{bojar2017findings}
Ond{\v{r}}ej Bojar, Rajen Chatterjee, Christian Federmann, Yvette Graham, Barry Haddow, Shujian Huang, Matthias Huck, Philipp Koehn, Qun Liu, Varvara Logacheva, Christof Monz, Matteo Negri, Matt Post, Raphael Rubino, Lucia Specia, and Marco Turchi.
\newblock Findings of the 2017 conference on machine translation ({WMT}17).
\newblock In Ond{\v{r}}ej Bojar, Christian Buck, Rajen Chatterjee, Christian Federmann, Yvette Graham, Barry Haddow, Matthias Huck, Antonio~Jimeno Yepes, Philipp Koehn, and Julia Kreutzer (eds.), \emph{Proceedings of the Second Conference on Machine Translation}, pp.\  169--214, Copenhagen, Denmark, September 2017. Association for Computational Linguistics.
\newblock \doi{10.18653/v1/W17-4717}.
\newblock URL \url{https://aclanthology.org/W17-4717}.

\bibitem[Bojar et~al.(2018)Bojar, Federmann, Fishel, Graham, Haddow, Huck, Koehn, and Monz]{bojar-etal-2018-findings}
Ond{\v{r}}ej Bojar, Christian Federmann, Mark Fishel, Yvette Graham, Barry Haddow, Matthias Huck, Philipp Koehn, and Christof Monz.
\newblock Findings of the 2018 conference on machine translation ({WMT}18).
\newblock In Ond{\v{r}}ej Bojar, Rajen Chatterjee, Christian Federmann, Mark Fishel, Yvette Graham, Barry Haddow, Matthias Huck, Antonio~Jimeno Yepes, Philipp Koehn, Christof Monz, Matteo Negri, Aur{\'e}lie N{\'e}v{\'e}ol, Mariana Neves, Matt Post, Lucia Specia, Marco Turchi, and Karin Verspoor (eds.), \emph{Proceedings of the Third Conference on Machine Translation: Shared Task Papers}, pp.\  272--303, Belgium, Brussels, October 2018. Association for Computational Linguistics.
\newblock \doi{10.18653/v1/W18-6401}.
\newblock URL \url{https://aclanthology.org/W18-6401}.

\bibitem[Cer et~al.(2017)Cer, Diab, Agirre, Lopez-Gazpio, and Specia]{cer-etal-2017-semeval}
Daniel Cer, Mona Diab, Eneko Agirre, I{\~n}igo Lopez-Gazpio, and Lucia Specia.
\newblock {S}em{E}val-2017 task 1: Semantic textual similarity multilingual and crosslingual focused evaluation.
\newblock In Steven Bethard, Marine Carpuat, Marianna Apidianaki, Saif~M. Mohammad, Daniel Cer, and David Jurgens (eds.), \emph{Proceedings of the 11th International Workshop on Semantic Evaluation ({S}em{E}val-2017)}, pp.\  1--14, Vancouver, Canada, August 2017. Association for Computational Linguistics.
\newblock \doi{10.18653/v1/S17-2001}.
\newblock URL \url{https://aclanthology.org/S17-2001}.

\bibitem[Cettolo et~al.(2014)Cettolo, Niehues, St{\"u}ker, Bentivogli, and Federico]{cettolo2014iwslt}
Mauro Cettolo, Jan Niehues, Sebastian St{\"u}ker, Luisa Bentivogli, and Marcello Federico.
\newblock Report on the 11th {IWSLT} evaluation campaign.
\newblock In Marcello Federico, Sebastian St{\"u}ker, and Fran{\c{c}}ois Yvon (eds.), \emph{Proceedings of the 11th International Workshop on Spoken Language Translation: Evaluation Campaign}, pp.\  2--17, Lake Tahoe, California, December 4-5 2014.
\newblock URL \url{https://aclanthology.org/2014.iwslt-evaluation.1}.

\bibitem[Cettolo et~al.(2017)Cettolo, Federico, Bentivogli, Niehues, St{\"u}ker, Sudoh, Yoshino, and Federmann]{cettolo-etal-2017-overview}
Mauro Cettolo, Marcello Federico, Luisa Bentivogli, Jan Niehues, Sebastian St{\"u}ker, Katsuhito Sudoh, Koichiro Yoshino, and Christian Federmann.
\newblock Overview of the {IWSLT} 2017 evaluation campaign.
\newblock In Sakriani Sakti and Masao Utiyama (eds.), \emph{Proceedings of the 14th International Conference on Spoken Language Translation}, pp.\  2--14, Tokyo, Japan, December 14-15 2017. International Workshop on Spoken Language Translation.
\newblock URL \url{https://aclanthology.org/2017.iwslt-1.1}.

\bibitem[Cheng \& Vlachos(2023)Cheng and Vlachos]{cheng-vlachos-2023-faster}
Julius Cheng and Andreas Vlachos.
\newblock Faster minimum {B}ayes risk decoding with confidence-based pruning.
\newblock In Houda Bouamor, Juan Pino, and Kalika Bali (eds.), \emph{Proceedings of the 2023 Conference on Empirical Methods in Natural Language Processing}, pp.\  12473--12480, Singapore, December 2023. Association for Computational Linguistics.
\newblock \doi{10.18653/v1/2023.emnlp-main.767}.
\newblock URL \url{https://aclanthology.org/2023.emnlp-main.767}.

\bibitem[Cohen \& Beck(2019)Cohen and Beck]{pmlr-v97-cohen19a}
Eldan Cohen and Christopher Beck.
\newblock Empirical analysis of beam search performance degradation in neural sequence models.
\newblock In Kamalika Chaudhuri and Ruslan Salakhutdinov (eds.), \emph{International Conference on Machine Learning}, volume~97 of \emph{Proceedings of Machine Learning Research}, pp.\  1290--1299. PMLR, 09--15 Jun 2019.
\newblock URL \url{https://proceedings.mlr.press/v97/cohen19a.html}.

\bibitem[Cong et~al.(2024)Cong, Daheim, Shen, Cremers, Yokota, Khan, and M{\"o}llenhoff]{cong2024variational}
Bai Cong, Nico Daheim, Yuesong Shen, Daniel Cremers, Rio Yokota, Mohammad~Emtiyaz Khan, and Thomas M{\"o}llenhoff.
\newblock Variational low-rank adaptation using {IVON}.
\newblock In \emph{NeurIPS 2024 Workshop on Fine-Tuning in Modern Machine Learning: Principles and Scalability}, 2024.
\newblock URL \url{https://openreview.net/forum?id=nRD5uZa2fe}.

\bibitem[Dale et~al.(2023)Dale, Voita, Barrault, and Costa-juss{\`a}]{dale-etal-2023-detecting}
David Dale, Elena Voita, Loic Barrault, and Marta~R. Costa-juss{\`a}.
\newblock Detecting and mitigating hallucinations in machine translation: Model internal workings alone do well, sentence similarity {E}ven better.
\newblock In Anna Rogers, Jordan Boyd-Graber, and Naoaki Okazaki (eds.), \emph{Proceedings of the 61st Annual Meeting of the Association for Computational Linguistics (Volume 1: Long Papers)}, pp.\  36--50, Toronto, Canada, July 2023. Association for Computational Linguistics.
\newblock \doi{10.18653/v1/2023.acl-long.3}.
\newblock URL \url{https://aclanthology.org/2023.acl-long.3}.

\bibitem[Daxberger et~al.(2021)Daxberger, Kristiadi, Immer, Eschenhagen, Bauer, and Hennig]{laplace2021}
Erik Daxberger, Agustinus Kristiadi, Alexander Immer, Runa Eschenhagen, Matthias Bauer, and Philipp Hennig.
\newblock Laplace redux--effortless {B}ayesian deep learning.
\newblock In \emph{{N}eur{IPS}}, 2021.
\newblock URL \url{https://proceedings.neurips.cc/paper_files/paper/2021/file/a7c9585703d275249f30a088cebba0ad-Paper.pdf}.

\bibitem[DeGroot(2005)]{degroot2005optimal}
Morris~H DeGroot.
\newblock \emph{Optimal statistical decisions}.
\newblock John Wiley \& Sons, 2005.
\newblock URL \url{https://onlinelibrary.wiley.com/doi/book/10.1002/0471729000}.

\bibitem[Dubey et~al.(2024)Dubey, Jauhri, Pandey, Kadian, Al-Dahle, Letman, Mathur, Schelten, Yang, Fan, et~al.]{dubey2024llama3herdmodels}
Abhimanyu Dubey, Abhinav Jauhri, Abhinav Pandey, Abhishek Kadian, Ahmad Al-Dahle, Aiesha Letman, Akhil Mathur, Alan Schelten, Amy Yang, Angela Fan, et~al.
\newblock The llama 3 herd of models.
\newblock \emph{arXiv preprint arXiv:2407.21783}, 2024.
\newblock URL \url{https://arxiv.org/abs/2407.21783}.

\bibitem[Eikema \& Aziz(2020)Eikema and Aziz]{eikema-aziz-2020-map}
Bryan Eikema and Wilker Aziz.
\newblock Is {MAP} decoding all you need? the inadequacy of the mode in neural machine translation.
\newblock In Donia Scott, Nuria Bel, and Chengqing Zong (eds.), \emph{Proceedings of the 28th International Conference on Computational Linguistics}, pp.\  4506--4520, Barcelona, Spain (Online), December 2020. International Committee on Computational Linguistics.
\newblock \doi{10.18653/v1/2020.coling-main.398}.
\newblock URL \url{https://aclanthology.org/2020.coling-main.398}.

\bibitem[Eikema \& Aziz(2022)Eikema and Aziz]{eikema-aziz-2022-sampling}
Bryan Eikema and Wilker Aziz.
\newblock Sampling-based approximations to minimum {B}ayes risk decoding for neural machine translation.
\newblock In Yoav Goldberg, Zornitsa Kozareva, and Yue Zhang (eds.), \emph{Proceedings of the 2022 Conference on Empirical Methods in Natural Language Processing}, pp.\  10978--10993, Abu Dhabi, United Arab Emirates, December 2022. Association for Computational Linguistics.
\newblock \doi{10.18653/v1/2022.emnlp-main.754}.
\newblock URL \url{https://aclanthology.org/2022.emnlp-main.754}.

\bibitem[Fadeeva et~al.(2024)Fadeeva, Rubashevskii, Shelmanov, Petrakov, Li, Mubarak, Tsymbalov, Kuzmin, Panchenko, Baldwin, Nakov, and Panov]{fadeeva-etal-2024-fact}
Ekaterina Fadeeva, Aleksandr Rubashevskii, Artem Shelmanov, Sergey Petrakov, Haonan Li, Hamdy Mubarak, Evgenii Tsymbalov, Gleb Kuzmin, Alexander Panchenko, Timothy Baldwin, Preslav Nakov, and Maxim Panov.
\newblock Fact-checking the output of large language models via token-level uncertainty quantification.
\newblock In Lun-Wei Ku, Andre Martins, and Vivek Srikumar (eds.), \emph{Findings of the Association for Computational Linguistics ACL 2024}, pp.\  9367--9385, Bangkok, Thailand and virtual meeting, August 2024. Association for Computational Linguistics.
\newblock URL \url{https://aclanthology.org/2024.findings-acl.558}.

\bibitem[Fan et~al.(2018)Fan, Lewis, and Dauphin]{fan-etal-2018-hierarchical}
Angela Fan, Mike Lewis, and Yann Dauphin.
\newblock Hierarchical neural story generation.
\newblock In Iryna Gurevych and Yusuke Miyao (eds.), \emph{Proceedings of the 56th Annual Meeting of the Association for Computational Linguistics (Volume 1: Long Papers)}, pp.\  889--898, Melbourne, Australia, July 2018. Association for Computational Linguistics.
\newblock \doi{10.18653/v1/P18-1082}.
\newblock URL \url{https://aclanthology.org/P18-1082}.

\bibitem[Farinhas et~al.(2023)Farinhas, de~Souza, and Martins]{farinhas-etal-2023-empirical}
Ant{\'o}nio Farinhas, Jos{\'e} de~Souza, and Andre Martins.
\newblock An empirical study of translation hypothesis ensembling with large language models.
\newblock In Houda Bouamor, Juan Pino, and Kalika Bali (eds.), \emph{Proceedings of the 2023 Conference on Empirical Methods in Natural Language Processing}, pp.\  11956--11970, Singapore, December 2023. Association for Computational Linguistics.
\newblock \doi{10.18653/v1/2023.emnlp-main.733}.
\newblock URL \url{https://aclanthology.org/2023.emnlp-main.733}.

\bibitem[Feng et~al.(2022)Feng, Yang, Cer, Arivazhagan, and Wang]{feng-etal-2022-language}
Fangxiaoyu Feng, Yinfei Yang, Daniel Cer, Naveen Arivazhagan, and Wei Wang.
\newblock Language-agnostic {BERT} sentence embedding.
\newblock In Smaranda Muresan, Preslav Nakov, and Aline Villavicencio (eds.), \emph{Proceedings of the 60th Annual Meeting of the Association for Computational Linguistics (Volume 1: Long Papers)}, pp.\  878--891, Dublin, Ireland, May 2022. Association for Computational Linguistics.
\newblock \doi{10.18653/v1/2022.acl-long.62}.
\newblock URL \url{https://aclanthology.org/2022.acl-long.62}.

\bibitem[Fernandes et~al.(2022)Fernandes, Farinhas, Rei, C.~de Souza, Ogayo, Neubig, and Martins]{fernandes-etal-2022-quality}
Patrick Fernandes, Ant{\'o}nio Farinhas, Ricardo Rei, Jos{\'e}~G. C.~de Souza, Perez Ogayo, Graham Neubig, and Andre Martins.
\newblock Quality-aware decoding for neural machine translation.
\newblock In Marine Carpuat, Marie-Catherine de~Marneffe, and Ivan~Vladimir Meza~Ruiz (eds.), \emph{Proceedings of the 2022 Conference of the North American Chapter of the Association for Computational Linguistics: Human Language Technologies}, pp.\  1396--1412, Seattle, United States, July 2022. Association for Computational Linguistics.
\newblock \doi{10.18653/v1/2022.naacl-main.100}.
\newblock URL \url{https://aclanthology.org/2022.naacl-main.100}.

\bibitem[Finkelstein \& Freitag(2024)Finkelstein and Freitag]{finkelstein2024mbr}
Mara Finkelstein and Markus Freitag.
\newblock {MBR} and {QE} finetuning: Training-time distillation of the best and most expensive decoding methods.
\newblock In \emph{The Twelfth International Conference on Learning Representations}, 2024.
\newblock URL \url{https://openreview.net/forum?id=bkNx3O0sND}.

\bibitem[Fort et~al.(2019)Fort, Hu, and Lakshminarayanan]{fort2019deep}
Stanislav Fort, Huiyi Hu, and Balaji Lakshminarayanan.
\newblock Deep ensembles: A loss landscape perspective.
\newblock \emph{arXiv preprint arXiv:1912.02757}, 2019.
\newblock URL \url{https://arxiv.org/abs/1912.02757}.

\bibitem[Frankle et~al.(2020)Frankle, Dziugaite, Roy, and Carbin]{pmlr-v119-frankle20a}
Jonathan Frankle, Gintare~Karolina Dziugaite, Daniel Roy, and Michael Carbin.
\newblock Linear mode connectivity and the lottery ticket hypothesis.
\newblock In Hal~Daumé III and Aarti Singh (eds.), \emph{Proceedings of the 37th International Conference on Machine Learning}, volume 119 of \emph{Proceedings of Machine Learning Research}, pp.\  3259--3269. PMLR, 13--18 Jul 2020.
\newblock URL \url{https://proceedings.mlr.press/v119/frankle20a.html}.

\bibitem[Freitag et~al.(2022)Freitag, Grangier, Tan, and Liang]{freitag-etal-2022-high}
Markus Freitag, David Grangier, Qijun Tan, and Bowen Liang.
\newblock High quality rather than high model probability: Minimum {B}ayes risk decoding with neural metrics.
\newblock \emph{Transactions of the Association for Computational Linguistics}, 10:\penalty0 811--825, 2022.
\newblock \doi{10.1162/tacl_a_00491}.
\newblock URL \url{https://aclanthology.org/2022.tacl-1.47}.

\bibitem[Geifman \& El-Yaniv(2017)Geifman and El-Yaniv]{geifman2017selective}
Yonatan Geifman and Ran El-Yaniv.
\newblock Selective classification for deep neural networks.
\newblock \emph{Advances in neural information processing systems}, 30, 2017.
\newblock URL \url{https://papers.nips.cc/paper_files/paper/2017/hash/4a8423d5e91fda00bb7e46540e2b0cf1-Abstract.html}.

\bibitem[{}{Gemma Team}(2024{\natexlab{a}})]{gemmateam2024gemma2improvingopen}
{}{Gemma Team}.
\newblock Gemma 2: Improving open language models at a practical size.
\newblock \emph{arXiv preprint arXiv:2408.00118}, 2024{\natexlab{a}}.
\newblock URL \url{https://arxiv.org/abs/2408.00118}.

\bibitem[{}{Gemma Team}(2024{\natexlab{b}})]{gemmateam2024gemmaopenmodelsbased}
{}{Gemma Team}.
\newblock Gemma: Open models based on gemini research and technology.
\newblock \emph{arXiv preprint arXiv:2403.08295}, 2024{\natexlab{b}}.
\newblock URL \url{https://arxiv.org/abs/2403.08295}.

\bibitem[Gliwa et~al.(2019)Gliwa, Mochol, Biesek, and Wawer]{gliwa-etal-2019-samsum}
Bogdan Gliwa, Iwona Mochol, Maciej Biesek, and Aleksander Wawer.
\newblock {SAMS}um corpus: A human-annotated dialogue dataset for abstractive summarization.
\newblock In Lu~Wang, Jackie Chi~Kit Cheung, Giuseppe Carenini, and Fei Liu (eds.), \emph{Proceedings of the 2nd Workshop on New Frontiers in Summarization}, pp.\  70--79, Hong Kong, China, November 2019. Association for Computational Linguistics.
\newblock \doi{10.18653/v1/D19-5409}.
\newblock URL \url{https://aclanthology.org/D19-5409}.

\bibitem[Gonz{\'a}lez-Rubio et~al.(2011)Gonz{\'a}lez-Rubio, Juan, and Casacuberta]{gonzalez-rubio-etal-2011-minimum}
Jes{\'u}s Gonz{\'a}lez-Rubio, Alfons Juan, and Francisco Casacuberta.
\newblock Minimum {B}ayes-risk system combination.
\newblock In Dekang Lin, Yuji Matsumoto, and Rada Mihalcea (eds.), \emph{Proceedings of the 49th Annual Meeting of the Association for Computational Linguistics: Human Language Technologies}, pp.\  1268--1277, Portland, Oregon, USA, June 2011. Association for Computational Linguistics.
\newblock URL \url{https://aclanthology.org/P11-1127}.

\bibitem[Graves(2011)]{graves2011practical}
Alex Graves.
\newblock Practical variational inference for neural networks.
\newblock \emph{Advances in neural information processing systems}, 24, 2011.
\newblock URL \url{https://papers.nips.cc/paper_files/paper/2011/hash/7eb3c8be3d411e8ebfab08eba5f49632-Abstract.html}.

\bibitem[Hewitt et~al.(2022)Hewitt, Manning, and Liang]{hewitt-etal-2022-truncation}
John Hewitt, Christopher Manning, and Percy Liang.
\newblock Truncation sampling as language model desmoothing.
\newblock In Yoav Goldberg, Zornitsa Kozareva, and Yue Zhang (eds.), \emph{Findings of the Association for Computational Linguistics: EMNLP 2022}, pp.\  3414--3427, Abu Dhabi, United Arab Emirates, December 2022. Association for Computational Linguistics.
\newblock \doi{10.18653/v1/2022.findings-emnlp.249}.
\newblock URL \url{https://aclanthology.org/2022.findings-emnlp.249}.

\bibitem[Himmi et~al.(2024)Himmi, Staerman, Picot, Colombo, and Guerreiro]{himmi2024enhanced}
Anas Himmi, Guillaume Staerman, Marine Picot, Pierre Colombo, and Nuno~M Guerreiro.
\newblock Enhanced hallucination detection in neural machine translation through simple detector aggregation.
\newblock In Yaser Al-Onaizan, Mohit Bansal, and Yun-Nung Chen (eds.), \emph{Proceedings of the 2024 Conference on Empirical Methods in Natural Language Processing}, pp.\  18573--18583, Miami, Florida, USA, November 2024. Association for Computational Linguistics.
\newblock \doi{10.18653/v1/2024.emnlp-main.1033}.
\newblock URL \url{https://aclanthology.org/2024.emnlp-main.1033/}.

\bibitem[Holtzman et~al.(2020)Holtzman, Buys, Du, Forbes, and Choi]{Holtzman2020The}
Ari Holtzman, Jan Buys, Li~Du, Maxwell Forbes, and Yejin Choi.
\newblock The curious case of neural text degeneration.
\newblock In \emph{International Conference on Learning Representations}, 2020.
\newblock URL \url{https://openreview.net/forum?id=rygGQyrFvH}.

\bibitem[Hu et~al.(2022)Hu, yelong shen, Wallis, Allen-Zhu, Li, Wang, Wang, and Chen]{hu2022lora}
Edward~J Hu, yelong shen, Phillip Wallis, Zeyuan Allen-Zhu, Yuanzhi Li, Shean Wang, Lu~Wang, and Weizhu Chen.
\newblock Lo{RA}: Low-rank adaptation of large language models.
\newblock In \emph{International Conference on Learning Representations}, 2022.
\newblock URL \url{https://openreview.net/forum?id=nZeVKeeFYf9}.

\bibitem[Huang et~al.(2017)Huang, Li, Pleiss, Liu, Hopcroft, and Weinberger]{huang2017snapshot}
Gao Huang, Yixuan Li, Geoff Pleiss, Zhuang Liu, John~E. Hopcroft, and Kilian~Q. Weinberger.
\newblock Snapshot ensembles: Train 1, get m for free.
\newblock In \emph{International Conference on Learning Representations}, 2017.
\newblock URL \url{https://openreview.net/forum?id=BJYwwY9ll}.

\bibitem[Jiang et~al.(2023)Jiang, Sablayrolles, Mensch, Bamford, Chaplot, de~las Casas, Bressand, Lengyel, Lample, Saulnier, Lavaud, Lachaux, Stock, Scao, Lavril, Wang, Lacroix, and Sayed]{jiang2023mistral7b}
Albert~Q. Jiang, Alexandre Sablayrolles, Arthur Mensch, Chris Bamford, Devendra~Singh Chaplot, Diego de~las Casas, Florian Bressand, Gianna Lengyel, Guillaume Lample, Lucile Saulnier, Lélio~Renard Lavaud, Marie-Anne Lachaux, Pierre Stock, Teven~Le Scao, Thibaut Lavril, Thomas Wang, Timothée Lacroix, and William~El Sayed.
\newblock Mistral 7b, 2023.
\newblock URL \url{https://arxiv.org/abs/2310.06825}.

\bibitem[Khan \& Rue(2023)Khan and Rue]{khan2023bayesian}
Mohammad~Emtiyaz Khan and H{\aa}vard Rue.
\newblock The bayesian learning rule.
\newblock \emph{Journal of Machine Learning Research}, 24\penalty0 (281):\penalty0 1--46, 2023.
\newblock URL \url{http://jmlr.org/papers/v24/22-0291.html}.

\bibitem[Kobayashi(2018)]{kobayashi-2018-frustratingly}
Hayato Kobayashi.
\newblock Frustratingly easy model ensemble for abstractive summarization.
\newblock In Ellen Riloff, David Chiang, Julia Hockenmaier, and Jun{'}ichi Tsujii (eds.), \emph{Proceedings of the 2018 Conference on Empirical Methods in Natural Language Processing}, pp.\  4165--4176, Brussels, Belgium, October-November 2018. Association for Computational Linguistics.
\newblock \doi{10.18653/v1/D18-1449}.
\newblock URL \url{https://aclanthology.org/D18-1449}.

\bibitem[Kocmi et~al.(2024)Kocmi, Zouhar, Federmann, and Post]{kocmi-etal-2024-navigating}
Tom Kocmi, Vil{\'e}m Zouhar, Christian Federmann, and Matt Post.
\newblock Navigating the metrics maze: Reconciling score magnitudes and accuracies.
\newblock In Lun-Wei Ku, Andre Martins, and Vivek Srikumar (eds.), \emph{Proceedings of the 62nd Annual Meeting of the Association for Computational Linguistics (Volume 1: Long Papers)}, pp.\  1999--2014, Bangkok, Thailand, August 2024. Association for Computational Linguistics.
\newblock URL \url{https://aclanthology.org/2024.acl-long.110}.

\bibitem[Kryscinski et~al.(2020)Kryscinski, McCann, Xiong, and Socher]{kryscinski-etal-2020-evaluating}
Wojciech Kryscinski, Bryan McCann, Caiming Xiong, and Richard Socher.
\newblock Evaluating the factual consistency of abstractive text summarization.
\newblock In Bonnie Webber, Trevor Cohn, Yulan He, and Yang Liu (eds.), \emph{Proceedings of the 2020 Conference on Empirical Methods in Natural Language Processing (EMNLP)}, pp.\  9332--9346, Online, November 2020. Association for Computational Linguistics.
\newblock \doi{10.18653/v1/2020.emnlp-main.750}.
\newblock URL \url{https://aclanthology.org/2020.emnlp-main.750}.

\bibitem[Kuhn et~al.(2023)Kuhn, Gal, and Farquhar]{kuhn2023semantic}
Lorenz Kuhn, Yarin Gal, and Sebastian Farquhar.
\newblock Semantic uncertainty: Linguistic invariances for uncertainty estimation in natural language generation.
\newblock In \emph{The Eleventh International Conference on Learning Representations}, 2023.
\newblock URL \url{https://openreview.net/forum?id=VD-AYtP0dve}.

\bibitem[Kumar \& Byrne(2002)Kumar and Byrne]{kumar-byrne-2002-minimum}
Shankar Kumar and William Byrne.
\newblock Minimum {B}ayes-risk word alignments of bilingual texts.
\newblock In \emph{Proceedings of the 2002 Conference on Empirical Methods in Natural Language Processing ({EMNLP} 2002)}, pp.\  140--147. Association for Computational Linguistics, July 2002.
\newblock \doi{10.3115/1118693.1118712}.
\newblock URL \url{https://aclanthology.org/W02-1019}.

\bibitem[Lakshminarayanan et~al.(2017)Lakshminarayanan, Pritzel, and Blundell]{lakshminarayanan2017simple}
Balaji Lakshminarayanan, Alexander Pritzel, and Charles Blundell.
\newblock Simple and scalable predictive uncertainty estimation using deep ensembles.
\newblock \emph{Advances in neural information processing systems}, 30, 2017.
\newblock URL \url{https://papers.nips.cc/paper_files/paper/2017/hash/9ef2ed4b7fd2c810847ffa5fa85bce38-Abstract.html}.

\bibitem[Li et~al.(2024)Li, Chen, Ren, Cheng, Zhao, Nie, and Wen]{li-etal-2024-dawn}
Junyi Li, Jie Chen, Ruiyang Ren, Xiaoxue Cheng, Xin Zhao, Jian-Yun Nie, and Ji-Rong Wen.
\newblock The dawn after the dark: An empirical study on factuality hallucination in large language models.
\newblock In Lun-Wei Ku, Andre Martins, and Vivek Srikumar (eds.), \emph{Proceedings of the 62nd Annual Meeting of the Association for Computational Linguistics (Volume 1: Long Papers)}, pp.\  10879--10899, Bangkok, Thailand, August 2024. Association for Computational Linguistics.
\newblock URL \url{https://aclanthology.org/2024.acl-long.586}.

\bibitem[Lin(2004)]{lin-2004-rouge}
Chin-Yew Lin.
\newblock {ROUGE}: A package for automatic evaluation of summaries.
\newblock In \emph{Text Summarization Branches Out}, pp.\  74--81, Barcelona, Spain, July 2004. Association for Computational Linguistics.
\newblock URL \url{https://aclanthology.org/W04-1013}.

\bibitem[Lion et~al.(2023)Lion, Bachmann, Noci, and Hofmann]{lion2023how}
Kai Lion, Gregor Bachmann, Lorenzo Noci, and Thomas Hofmann.
\newblock How good is a single basin?
\newblock In \emph{UniReps: the First Workshop on Unifying Representations in Neural Models}, 2023.
\newblock URL \url{https://openreview.net/forum?id=JYww68b9PA}.

\bibitem[Loshchilov \& Hutter(2019)Loshchilov and Hutter]{loshchilov2018decoupled}
Ilya Loshchilov and Frank Hutter.
\newblock Decoupled weight decay regularization.
\newblock In \emph{International Conference on Learning Representations}, 2019.
\newblock URL \url{https://openreview.net/forum?id=Bkg6RiCqY7}.

\bibitem[Lukasik et~al.(2024)Lukasik, Narasimhan, Menon, Yu, and Kumar]{lukasik2024metricawarellminferenceregression}
Michal Lukasik, Harikrishna Narasimhan, Aditya~Krishna Menon, Felix Yu, and Sanjiv Kumar.
\newblock Regression aware inference with {LLM}s.
\newblock In Yaser Al-Onaizan, Mohit Bansal, and Yun-Nung Chen (eds.), \emph{Findings of the Association for Computational Linguistics: EMNLP 2024}, pp.\  13667--13678, Miami, Florida, USA, November 2024. Association for Computational Linguistics.
\newblock \doi{10.18653/v1/2024.findings-emnlp.799}.
\newblock URL \url{https://aclanthology.org/2024.findings-emnlp.799/}.

\bibitem[Mackay(1992)]{mackay1992bayesian}
David John~Cameron Mackay.
\newblock \emph{Bayesian methods for adaptive models}.
\newblock California Institute of Technology, 1992.
\newblock URL \url{https://thesis.library.caltech.edu/25/}.

\bibitem[Maddox et~al.(2019)Maddox, Izmailov, Garipov, Vetrov, and Wilson]{maddox2019simple}
Wesley~J Maddox, Pavel Izmailov, Timur Garipov, Dmitry~P Vetrov, and Andrew~Gordon Wilson.
\newblock A simple baseline for bayesian uncertainty in deep learning.
\newblock \emph{Advances in neural information processing systems}, 32, 2019.
\newblock URL \url{https://papers.nips.cc/paper_files/paper/2019/hash/118921efba23fc329e6560b27861f0c2-Abstract.html}.

\bibitem[Malinin \& Gales(2021)Malinin and Gales]{malinin2021uncertainty}
Andrey Malinin and Mark Gales.
\newblock Uncertainty estimation in autoregressive structured prediction.
\newblock In \emph{International Conference on Learning Representations}, 2021.
\newblock URL \url{https://openreview.net/forum?id=jN5y-zb5Q7m}.

\bibitem[Masegosa(2020)]{masegosa2020learning}
Andres Masegosa.
\newblock Learning under model misspecification: Applications to variational and ensemble methods.
\newblock \emph{Advances in Neural Information Processing Systems}, 33:\penalty0 5479--5491, 2020.
\newblock URL \url{https://proceedings.neurips.cc/paper/2020/hash/3ac48664b7886cf4e4ab4aba7e6b6bc9-Abstract.html}.

\bibitem[M{\"o}llenhoff \& Khan(2023)M{\"o}llenhoff and Khan]{mollenhoff2023sam}
Thomas M{\"o}llenhoff and Mohammad~Emtiyaz Khan.
\newblock {SAM} as an optimal relaxation of bayes.
\newblock In \emph{The Eleventh International Conference on Learning Representations}, 2023.
\newblock URL \url{https://openreview.net/forum?id=k4fevFqSQcX}.

\bibitem[Narayan et~al.(2018)Narayan, Cohen, and Lapata]{narayan-etal-2018-dont}
Shashi Narayan, Shay~B. Cohen, and Mirella Lapata.
\newblock Don{'}t give me the details, just the summary! topic-aware convolutional neural networks for extreme summarization.
\newblock In Ellen Riloff, David Chiang, Julia Hockenmaier, and Jun{'}ichi Tsujii (eds.), \emph{Proceedings of the 2018 Conference on Empirical Methods in Natural Language Processing}, pp.\  1797--1807, Brussels, Belgium, October-November 2018. Association for Computational Linguistics.
\newblock \doi{10.18653/v1/D18-1206}.
\newblock URL \url{https://aclanthology.org/D18-1206}.

\bibitem[Novikova et~al.(2017)Novikova, Du{\v{s}}ek, and Rieser]{novikova2017e2e}
Jekaterina Novikova, Ondrej Du{\v{s}}ek, and Verena Rieser.
\newblock The {E2E} dataset: New challenges for end-to-end generation.
\newblock In \emph{Proceedings of the 18th Annual Meeting of the Special Interest Group on Discourse and Dialogue}, Saarbr\"ucken, Germany, 2017.
\newblock URL \url{https://arxiv.org/abs/1706.09254}.
\newblock arXiv:1706.09254.

\bibitem[Osawa et~al.(2019)Osawa, Swaroop, Khan, Jain, Eschenhagen, Turner, and Yokota]{osawa2019practical}
Kazuki Osawa, Siddharth Swaroop, Mohammad Emtiyaz~E Khan, Anirudh Jain, Runa Eschenhagen, Richard~E Turner, and Rio Yokota.
\newblock Practical deep learning with bayesian principles.
\newblock \emph{Advances in neural information processing systems}, 32, 2019.
\newblock URL \url{https://papers.nips.cc/paper_files/paper/2019/hash/b53477c2821c1bf0da5d40e57b870d35-Abstract.html}.

\bibitem[Ott et~al.(2019)Ott, Edunov, Baevski, Fan, Gross, Ng, Grangier, and Auli]{ott-etal-2019-fairseq}
Myle Ott, Sergey Edunov, Alexei Baevski, Angela Fan, Sam Gross, Nathan Ng, David Grangier, and Michael Auli.
\newblock fairseq: A fast, extensible toolkit for sequence modeling.
\newblock In Waleed Ammar, Annie Louis, and Nasrin Mostafazadeh (eds.), \emph{Proceedings of the 2019 Conference of the North {A}merican Chapter of the Association for Computational Linguistics (Demonstrations)}, pp.\  48--53, Minneapolis, Minnesota, June 2019. Association for Computational Linguistics.
\newblock \doi{10.18653/v1/N19-4009}.
\newblock URL \url{https://aclanthology.org/N19-4009}.

\bibitem[Papineni et~al.(2002)Papineni, Roukos, Ward, and Zhu]{papineni-etal-2002-bleu}
Kishore Papineni, Salim Roukos, Todd Ward, and Wei-Jing Zhu.
\newblock {B}leu: a method for automatic evaluation of machine translation.
\newblock In Pierre Isabelle, Eugene Charniak, and Dekang Lin (eds.), \emph{Proceedings of the 40th Annual Meeting of the Association for Computational Linguistics}, pp.\  311--318, Philadelphia, Pennsylvania, USA, July 2002. Association for Computational Linguistics.
\newblock \doi{10.3115/1073083.1073135}.
\newblock URL \url{https://aclanthology.org/P02-1040}.

\bibitem[Popovi{\'c}(2015)]{popovic-2015-chrf}
Maja Popovi{\'c}.
\newblock chr{F}: character n-gram {F}-score for automatic {MT} evaluation.
\newblock In Ond{\v{r}}ej Bojar, Rajan Chatterjee, Christian Federmann, Barry Haddow, Chris Hokamp, Matthias Huck, Varvara Logacheva, and Pavel Pecina (eds.), \emph{Proceedings of the Tenth Workshop on Statistical Machine Translation}, pp.\  392--395, Lisbon, Portugal, September 2015. Association for Computational Linguistics.
\newblock \doi{10.18653/v1/W15-3049}.
\newblock URL \url{https://aclanthology.org/W15-3049}.

\bibitem[Post(2018)]{post-2018-call}
Matt Post.
\newblock A call for clarity in reporting {BLEU} scores.
\newblock In Ond{\v{r}}ej Bojar, Rajen Chatterjee, Christian Federmann, Mark Fishel, Yvette Graham, Barry Haddow, Matthias Huck, Antonio~Jimeno Yepes, Philipp Koehn, Christof Monz, Matteo Negri, Aur{\'e}lie N{\'e}v{\'e}ol, Mariana Neves, Matt Post, Lucia Specia, Marco Turchi, and Karin Verspoor (eds.), \emph{Proceedings of the Third Conference on Machine Translation: Research Papers}, pp.\  186--191, Brussels, Belgium, October 2018. Association for Computational Linguistics.
\newblock \doi{10.18653/v1/W18-6319}.
\newblock URL \url{https://aclanthology.org/W18-6319}.

\bibitem[Rei et~al.(2022)Rei, C.~de Souza, Alves, Zerva, Farinha, Glushkova, Lavie, Coheur, and Martins]{rei-etal-2022-comet}
Ricardo Rei, Jos{\'e}~G. C.~de Souza, Duarte Alves, Chrysoula Zerva, Ana~C Farinha, Taisiya Glushkova, Alon Lavie, Luisa Coheur, and Andr{\'e} F.~T. Martins.
\newblock {COMET}-22: Unbabel-{IST} 2022 submission for the metrics shared task.
\newblock In Philipp Koehn, Lo{\"\i}c Barrault, Ond{\v{r}}ej Bojar, Fethi Bougares, Rajen Chatterjee, Marta~R. Costa-juss{\`a}, Christian Federmann, Mark Fishel, Alexander Fraser, Markus Freitag, Yvette Graham, Roman Grundkiewicz, Paco Guzman, Barry Haddow, Matthias Huck, Antonio Jimeno~Yepes, Tom Kocmi, Andr{\'e} Martins, Makoto Morishita, Christof Monz, Masaaki Nagata, Toshiaki Nakazawa, Matteo Negri, Aur{\'e}lie N{\'e}v{\'e}ol, Mariana Neves, Martin Popel, Marco Turchi, and Marcos Zampieri (eds.), \emph{Proceedings of the Seventh Conference on Machine Translation (WMT)}, pp.\  578--585, Abu Dhabi, United Arab Emirates (Hybrid), December 2022. Association for Computational Linguistics.
\newblock URL \url{https://aclanthology.org/2022.wmt-1.52}.

\bibitem[Reid et~al.(2021)Reid, Hu, Neubig, and Matsuo]{reid-etal-2021-afromt}
Machel Reid, Junjie Hu, Graham Neubig, and Yutaka Matsuo.
\newblock {A}fro{MT}: Pretraining strategies and reproducible benchmarks for translation of 8 {A}frican languages.
\newblock In Marie-Francine Moens, Xuanjing Huang, Lucia Specia, and Scott Wen-tau Yih (eds.), \emph{Proceedings of the 2021 Conference on Empirical Methods in Natural Language Processing}, pp.\  1306--1320, Online and Punta Cana, Dominican Republic, November 2021. Association for Computational Linguistics.
\newblock \doi{10.18653/v1/2021.emnlp-main.99}.
\newblock URL \url{https://aclanthology.org/2021.emnlp-main.99}.

\bibitem[Reimers \& Gurevych(2019)Reimers and Gurevych]{reimers-2019-sentence-bert}
Nils Reimers and Iryna Gurevych.
\newblock Sentence-{BERT}: Sentence embeddings using {S}iamese {BERT}-networks.
\newblock In Kentaro Inui, Jing Jiang, Vincent Ng, and Xiaojun Wan (eds.), \emph{Proceedings of the 2019 Conference on Empirical Methods in Natural Language Processing and the 9th International Joint Conference on Natural Language Processing (EMNLP-IJCNLP)}, pp.\  3982--3992, Hong Kong, China, November 2019. Association for Computational Linguistics.
\newblock \doi{10.18653/v1/D19-1410}.
\newblock URL \url{https://aclanthology.org/D19-1410/}.

\bibitem[Ren et~al.(2023)Ren, Luo, Zhao, Krishna, Saleh, Lakshminarayanan, and Liu]{ren2023outofdistribution}
Jie Ren, Jiaming Luo, Yao Zhao, Kundan Krishna, Mohammad Saleh, Balaji Lakshminarayanan, and Peter~J Liu.
\newblock Out-of-distribution detection and selective generation for conditional language models.
\newblock In \emph{International Conference on Learning Representations}, 2023.
\newblock URL \url{https://openreview.net/forum?id=kJUS5nD0vPB}.

\bibitem[Robert(2007)]{robert2007bayesian}
Christian~P. Robert.
\newblock \emph{The Bayesian choice: from decision-theoretic foundations to computational implementation}, volume~2.
\newblock Springer, 2007.
\newblock URL \url{https://doi.org/10.1007/0-387-71599-1}.

\bibitem[Sennrich et~al.(2016)Sennrich, Haddow, and Birch]{sennrich-etal-2016-neural}
Rico Sennrich, Barry Haddow, and Alexandra Birch.
\newblock Neural machine translation of rare words with subword units.
\newblock In Katrin Erk and Noah~A. Smith (eds.), \emph{Proceedings of the 54th Annual Meeting of the Association for Computational Linguistics (Volume 1: Long Papers)}, pp.\  1715--1725, Berlin, Germany, August 2016. Association for Computational Linguistics.
\newblock \doi{10.18653/v1/P16-1162}.
\newblock URL \url{https://aclanthology.org/P16-1162}.

\bibitem[Shen et~al.(2019)Shen, Ott, Auli, and Ranzato]{shen2019mixture}
Tianxiao Shen, Myle Ott, Michael Auli, and Marc’Aurelio Ranzato.
\newblock Mixture models for diverse machine translation: Tricks of the trade.
\newblock In \emph{International conference on machine learning}, 2019.
\newblock URL \url{https://proceedings.mlr.press/v97/shen19c.html}.

\bibitem[Shen et~al.(2024)Shen, Daheim, Cong, Nickl, Marconi, Bazan, Yokota, Gurevych, Cremers, Khan, and Moellenhoff]{shen2024variational}
Yuesong Shen, Nico Daheim, Bai Cong, Peter Nickl, Gian~Maria Marconi, Clement Bazan, Rio Yokota, Iryna Gurevych, Daniel Cremers, Mohammad~Emtiyaz Khan, and Thomas Moellenhoff.
\newblock Variational learning is effective for large deep networks.
\newblock \emph{ICML}, 2024.
\newblock URL \url{https://openreview.net/forum?id=cXBv07GKvk}.

\bibitem[Smith(2011)]{smith:2011:synthesis}
Noah~A. Smith.
\newblock \emph{Linguistic Structure Prediction}.
\newblock Synthesis Lectures on Human Language Technologies. Morgan and Claypool, May 2011.
\newblock URL \url{https://link.springer.com/book/10.1007/978-3-031-02143-5}.

\bibitem[Stahlberg \& Byrne(2019)Stahlberg and Byrne]{stahlberg-byrne-2019-nmt}
Felix Stahlberg and Bill Byrne.
\newblock On {NMT} search errors and model errors: Cat got your tongue?
\newblock In Kentaro Inui, Jing Jiang, Vincent Ng, and Xiaojun Wan (eds.), \emph{Proceedings of the 2019 Conference on Empirical Methods in Natural Language Processing and the 9th International Joint Conference on Natural Language Processing (EMNLP-IJCNLP)}, pp.\  3356--3362, Hong Kong, China, November 2019. Association for Computational Linguistics.
\newblock \doi{10.18653/v1/D19-1331}.
\newblock URL \url{https://aclanthology.org/D19-1331}.

\bibitem[Suzgun et~al.(2023)Suzgun, Melas-Kyriazi, and Jurafsky]{suzgun-etal-2023-follow}
Mirac Suzgun, Luke Melas-Kyriazi, and Dan Jurafsky.
\newblock Follow the wisdom of the crowd: Effective text generation via minimum {B}ayes risk decoding.
\newblock In Anna Rogers, Jordan Boyd-Graber, and Naoaki Okazaki (eds.), \emph{Findings of the Association for Computational Linguistics: ACL 2023}, pp.\  4265--4293, Toronto, Canada, July 2023. Association for Computational Linguistics.
\newblock \doi{10.18653/v1/2023.findings-acl.262}.
\newblock URL \url{https://aclanthology.org/2023.findings-acl.262}.

\bibitem[Vamvas \& Sennrich(2024)Vamvas and Sennrich]{vamvas-sennrich-2024-linear}
Jannis Vamvas and Rico Sennrich.
\newblock Linear-time minimum {B}ayes risk decoding with reference aggregation.
\newblock In Lun-Wei Ku, Andre Martins, and Vivek Srikumar (eds.), \emph{Proceedings of the 62nd Annual Meeting of the Association for Computational Linguistics (Volume 2: Short Papers)}, pp.\  790--801, Bangkok, Thailand, August 2024. Association for Computational Linguistics.
\newblock \doi{10.18653/v1/2024.acl-short.71}.
\newblock URL \url{https://aclanthology.org/2024.acl-short.71}.

\bibitem[van~der Poel et~al.(2022)van~der Poel, Cotterell, and Meister]{van-der-poel-etal-2022-mutual}
Liam van~der Poel, Ryan Cotterell, and Clara Meister.
\newblock Mutual information alleviates hallucinations in abstractive summarization.
\newblock In Yoav Goldberg, Zornitsa Kozareva, and Yue Zhang (eds.), \emph{Proceedings of the 2022 Conference on Empirical Methods in Natural Language Processing}, pp.\  5956--5965, Abu Dhabi, United Arab Emirates, December 2022. Association for Computational Linguistics.
\newblock \doi{10.18653/v1/2022.emnlp-main.399}.
\newblock URL \url{https://aclanthology.org/2022.emnlp-main.399}.

\bibitem[Vaswani et~al.(2017)Vaswani, Shazeer, Parmar, Uszkoreit, Jones, Gomez, Kaiser, and Polosukhin]{vaswani2017}
Ashish Vaswani, Noam Shazeer, Niki Parmar, Jakob Uszkoreit, Llion Jones, Aidan~N Gomez, \L~ukasz Kaiser, and Illia Polosukhin.
\newblock Attention is all you need.
\newblock In I.~Guyon, U.~Von Luxburg, S.~Bengio, H.~Wallach, R.~Fergus, S.~Vishwanathan, and R.~Garnett (eds.), \emph{Advances in Neural Information Processing Systems}, volume~30. Curran Associates, Inc., 2017.
\newblock URL \url{https://proceedings.neurips.cc/paper_files/paper/2017/file/3f5ee243547dee91fbd053c1c4a845aa-Paper.pdf}.

\bibitem[Wenzel et~al.(2020)Wenzel, Roth, Veeling, Swiatkowski, Tran, Mandt, Snoek, Salimans, Jenatton, and Nowozin]{pmlr-v119-wenzel20a}
Florian Wenzel, Kevin Roth, Bastiaan Veeling, Jakub Swiatkowski, Linh Tran, Stephan Mandt, Jasper Snoek, Tim Salimans, Rodolphe Jenatton, and Sebastian Nowozin.
\newblock How good is the {B}ayes posterior in deep neural networks really?
\newblock In Hal~Daumé III and Aarti Singh (eds.), \emph{Proceedings of the 37th International Conference on Machine Learning}, volume 119 of \emph{Proceedings of Machine Learning Research}, pp.\  10248--10259. PMLR, 13--18 Jul 2020.
\newblock URL \url{https://proceedings.mlr.press/v119/wenzel20a.html}.

\bibitem[Wolf et~al.(2020)Wolf, Debut, Sanh, Chaumond, Delangue, Moi, Cistac, Rault, Louf, Funtowicz, Davison, Shleifer, von Platen, Ma, Jernite, Plu, Xu, Le~Scao, Gugger, Drame, Lhoest, and Rush]{wolf-etal-2020-transformers}
Thomas Wolf, Lysandre Debut, Victor Sanh, Julien Chaumond, Clement Delangue, Anthony Moi, Pierric Cistac, Tim Rault, Remi Louf, Morgan Funtowicz, Joe Davison, Sam Shleifer, Patrick von Platen, Clara Ma, Yacine Jernite, Julien Plu, Canwen Xu, Teven Le~Scao, Sylvain Gugger, Mariama Drame, Quentin Lhoest, and Alexander Rush.
\newblock Transformers: State-of-the-art natural language processing.
\newblock In Qun Liu and David Schlangen (eds.), \emph{Proceedings of the 2020 Conference on Empirical Methods in Natural Language Processing: System Demonstrations}, pp.\  38--45, Online, October 2020. Association for Computational Linguistics.
\newblock \doi{10.18653/v1/2020.emnlp-demos.6}.
\newblock URL \url{https://aclanthology.org/2020.emnlp-demos.6}.

\bibitem[Wood et~al.(2023)Wood, Mu, Webb, Reeve, Lujan, and Brown]{wood2023unified}
Danny Wood, Tingting Mu, Andrew~M Webb, Henry~WJ Reeve, Mikel Lujan, and Gavin Brown.
\newblock A unified theory of diversity in ensemble learning.
\newblock \emph{Journal of Machine Learning Research}, 24\penalty0 (359):\penalty0 1--49, 2023.
\newblock URL \url{https://jmlr.org/papers/v24/23-0041.html}.

\bibitem[Wu et~al.(2025)Wu, Fernandes, Bertsch, Kim, Pakazad, and Neubig]{wu2025better}
Ian Wu, Patrick Fernandes, Amanda Bertsch, Seungone Kim, Sina~Khoshfetrat Pakazad, and Graham Neubig.
\newblock Better instruction-following through minimum bayes risk.
\newblock In \emph{The Thirteenth International Conference on Learning Representations}, 2025.
\newblock URL \url{https://openreview.net/forum?id=7xCSK9BLPy}.

\bibitem[Xiao \& Wang(2021)Xiao and Wang]{xiao-wang-2021-hallucination}
Yijun Xiao and William~Yang Wang.
\newblock On hallucination and predictive uncertainty in conditional language generation.
\newblock In Paola Merlo, Jorg Tiedemann, and Reut Tsarfaty (eds.), \emph{Proceedings of the 16th Conference of the European Chapter of the Association for Computational Linguistics: Main Volume}, pp.\  2734--2744, Online, April 2021. Association for Computational Linguistics.
\newblock \doi{10.18653/v1/2021.eacl-main.236}.
\newblock URL \url{https://aclanthology.org/2021.eacl-main.236}.

\bibitem[Yang et~al.(2024{\natexlab{a}})Yang, Robeyns, Wang, and Aitchison]{yang2024bayesian}
Adam~X. Yang, Maxime Robeyns, Xi~Wang, and Laurence Aitchison.
\newblock Bayesian low-rank adaptation for large language models.
\newblock In \emph{The Twelfth International Conference on Learning Representations}, 2024{\natexlab{a}}.
\newblock URL \url{https://openreview.net/forum?id=FJiUyzOF1m}.

\bibitem[Yang et~al.(2024{\natexlab{b}})Yang, Chen, Lin, and Byrne]{yang-etal-2024-direct}
Guangyu Yang, Jinghong Chen, Weizhe Lin, and Bill Byrne.
\newblock Direct preference optimization for neural machine translation with minimum {B}ayes risk decoding.
\newblock In Kevin Duh, Helena Gomez, and Steven Bethard (eds.), \emph{Proceedings of the 2024 Conference of the North American Chapter of the Association for Computational Linguistics: Human Language Technologies (Volume 2: Short Papers)}, pp.\  391--398, Mexico City, Mexico, June 2024{\natexlab{b}}. Association for Computational Linguistics.
\newblock \doi{10.18653/v1/2024.naacl-short.34}.
\newblock URL \url{https://aclanthology.org/2024.naacl-short.34}.

\bibitem[{Yang et al.}(2024)]{yang2024qwen2technicalreport}
{An} {Yang et al.}
\newblock Qwen2 technical report, 2024.
\newblock URL \url{https://arxiv.org/abs/2407.10671}.

\bibitem[Ye et~al.(2023)Ye, Liu, Zhang, Hua, and Jia]{ye2023cognitivemiragereviewhallucinations}
Hongbin Ye, Tong Liu, Aijia Zhang, Wei Hua, and Weiqiang Jia.
\newblock Cognitive mirage: A review of hallucinations in large language models.
\newblock \emph{arXiv preprint arXiv:2309.06794}, 2023.
\newblock URL \url{https://arxiv.org/abs/2309.06794}.

\bibitem[Zhang et~al.(2020)Zhang, Kishore, Wu, Weinberger, and Artzi]{Zhang2020BERTScore}
Tianyi Zhang, Varsha Kishore, Felix Wu, Kilian~Q. Weinberger, and Yoav Artzi.
\newblock Bertscore: Evaluating text generation with bert.
\newblock In \emph{International Conference on Learning Representations}, 2020.
\newblock URL \url{https://openreview.net/forum?id=SkeHuCVFDr}.

\end{thebibliography}
\bibliographystyle{iclr2024_conference}

\appendix
\section{Experimental Details}
\label{app:experimental_details}

\subsection{Training from Scratch}
\label{app:from_scratch}

\paragraph{Datasets}
Our usage of the WMT14 English-to-German translation tasks~\citep{bojar14wmt} follows the set-up from~\citep{vaswani2017} but augments the training data by the \emph{news-commentary-v12} data from WMT17~\citep{bojar2017findings}.
In total, we train on ca. 3.9M paired examples.
We also use a validation set during training in order to pick checkpoints which consists of ca 39.4K examples.
We use the original \emph{newstest2014} data which consists of 3,003 examples for evaluation.

We also use the IWSLT14 German-to-English translation task~\citep{cettolo2014iwslt} which consists of ca 160K training examples.
The validation set consists of ca. 7.3K examples.
The test set consists of 6,750K examples.

Furthermore, we use two language pairs from AfroMT~\citep{reid-etal-2021-afromt}, namely En-Bem (English-Bemba) which consists of 275K training, 3K validation, and 3K test examples.
We do not use any monolingual data but only train from scratch on the parallel data.
We use En-Run (English-Rundi) in the same way, which consists of 253K training, 3k validation, and 3k test examples.

All data usages can be reproduced by following the instructions from the Fairseq repository under \url{https://github.com/facebookresearch/fairseq/tree/main/examples/translation} and will be published along our code.

\paragraph{Models}
All models follow the Transformer architecture from~\citet{vaswani2017} which consists of an encoder-decoder Transformer with 6 encoder and 6 decoder layers.
We use the Transformer$_\text{base}$ architecture for IWSLT2014 and afroMT and Transformer$_\text{big}$ for WMT14 which has larger embedding and feed forward dimensions.
The models use a vocabulary of Byte-Pair-Encoding tokens~\citep{sennrich-etal-2016-neural}.
The input and output embedding parameters of the decoder are shared.
The IWSLT model has an input vocabulary size of 8848 and an output vocabulary size of 6632 for in total $39,469,056$ parameters.
The en-run and en-bem models both have an input and output vocabulary size of 80000 each and a total of $126,058,496$ parameters.
The WMT model has an input vocabulary size of $40480$ and an output vocabulary size of $42720$ for a total of $261,431,296$ parameters.

\paragraph{Training \& Decoding}
We train all models from scratch using the fairseq library~\citep{ott-etal-2019-fairseq} which we extend for variational learning and a Bayesian interpretation of neural networks.
Fairseq is licensed under MIT license\footnote{\url{https://github.com/facebookresearch/fairseq/blob/main/LICENSE}} which permits our form of usage.
We will release our code publicly in the future for further research in a software repository under Apache License 2.0\footnote{\url{https://www.apache.org/licenses/LICENSE-2.0}}.
We train all models with the IVON optimizer~\citep{shen2024variational} and place a diagonal Gaussian posterior over neural networks.
We use IVON with a isotropic Gaussian prior and initialize all entries of the Hessian with $0.1$.
We use an effective sample size of $1\cdot10^{-8}$, a small weight-decay of $0.0001$, and a learning rate of $0.1$.
We set $\beta_1=0.9$ and $\beta_2=0.9999$.
All models are trained with a batch size of $32$ or up to $1024$ tokens and we use $2$ MC samples from the posterior during training for afroMT and IWSLT2014.
For WMT14 we just use one MC sample due to the heavier compute requirements.
We clip gradients elementwise at $0.001$ and use a dropout rate of $0.2$.
We train the models until performance in terms of BLEU has not improved for at least 3 epochs and then stop with the exception for WMT14, where we train only up to 20 epochs.
The results for the single model baseline and unimodal posterior are averaged over four runs.

For the snapshot-like approach, we add 3 randomly-sampled distributions that were trained with at least 10 epochs to the best-performing one.
For Deep Ensembles we always use four runs with different random seeds unless stated otherwise and for unimodal posteriors we sample four models from each posterior.
In all experiments we sample from the posterior ``as-is'' and only vary the temperature by reducing the effective sample size when explicitly mentioned.

All models are trained on a single GPU which is an NVIDIA GPU with either 80GB, 40GB, 32GB or 24GB GPU memory.
Training takes around 1-3 hours for the IWSLT14 and afroMT models and >2 days for the WMT models.

Following prior work, we use a length-penalty of $0.6$ for decoding~\citep{vaswani2017}. 

\subsection{Finetuning}
\label{app:finetuning}

\paragraph{Datasets}
For all datasets we use the versions from the huggingface hub (\url{https://huggingface.co/}).
We use the En-De split of the IWSLT17 evaluation campaign (\url{https://huggingface.co/datasets/IWSLT/iwslt2017})~\citep{cettolo-etal-2017-overview} with 206,122 training and 8079 test examples and the WMT18 Tr-En split (\url{https://huggingface.co/datasets/wmt/wmt18})~\citep{bojar-etal-2018-findings} with 205,756 training and 3,000 test examples for machine translation.
For summarization experiments, we use XSUM (\url{https://huggingface.co/datasets/EdinburghNLP/xsum})~\citep{narayan-etal-2018-dont} and SAMSum (\url{https://huggingface.co/datasets/Samsung/samsum})~\citep{gliwa-etal-2019-samsum}.
XSUM has 204,045 training examples---we train only on the first $50\%$ to reduce computational load---and 11,334 test examples.
SAMSum is much smaller and consists only of 14,732 train and 819 test examples.
Finally, we use E2E-NLG (\url{https://huggingface.co/datasets/tuetschek/e2e_nlg})~\citep{novikova2017e2e} with 33,524 train and 1,846 test examples for data-to-text generation, as well as STS-B (\url{https://huggingface.co/datasets/sentence-transformers/stsb})~\citep{cer-etal-2017-semeval} with 5,749 train and 1,379 test examples for sentence similarity scoring.
Note that we use the version provided with the sentence transformers library~\citep{reimers-2019-sentence-bert} which uses ratings from 0 to 1.

\paragraph{Models}
For finetuning results, we use the Gemma-2B-it~\citep{gemmateam2024gemmaopenmodelsbased} checkpoint, which can be found under \url{https://huggingface.co/google/gemma-2b-it} on the huggingface hub, with in total 2.51B parameters.

\paragraph{Training \& Decoding}
\begin{table*}[t]
    \centering
    \resizebox{\linewidth}{!}{\begin{tabular}{ll}
         Dataset & Instruction \\
    \hline
         IWSLT17 En-De & Translate from English to German: \\ 
         WMT18 Tr-En & Translate from Turkish to English: \\ 
         XSUM & Summarize: \\
         SamSum & Summarize: \\
         E2E-NLG & Convert a set of two-to-nine key-value attribute pairs in the restaurant domain to a simple English-language text: \\
         STSB & How similar are these sentences from 0 to 1? \\
         \hline
    \end{tabular}}
    \caption{Simple instructions used when finetuning Gemma-2B-it.}
    \label{tab:prompts}
\end{table*}
We finetune the model using LoRA~\citep{hu2022lora} with a rank $r=8$, $\alpha=32$ and a dropout rate of $0.1$.
In total, this introduces $921,600$ new parameters that are learned with IVON and, correspondingly, the diagonal variance consists of $921,600$ further parameters that are learned.
We use the chat template provided with huggingface~\citep{wolf-etal-2020-transformers}, which we adapt to organize our experiments in line with the Apache 2.0 license it is distributed under, to organize training and decoding.
As we use an instruction-tuned model, we use simple instructions for each dataset which are outlined in~\cref{tab:prompts}.
We train the model on both the prompt and the output labels and do not only calculate gradients for the latter.

We again use IVON to learn a unimodal diagonal Gaussian posterior. We use four separate runs with different random seeds for the Deep Ensembles (which entails different data order and initialization of new parameters) and sample four models for the unimodal posterior.
Results for the unimodal posterior and single model baseline are averaged over four seeds.
For all experiments we use the same hyperparameter setting.
We use an initial learning rate of $0.03$ which we anneal to $0$ with a cosine decay. We set $\beta_1 = 0.9$, $\beta_2 = 0.99999$, and use a small weight decay of $10^{-6}$.
We again clip gradients to unit norm and element-wise with a maximum value of $0.001$.
All hessian values are initialized at 0.0003.
We set the effective sample size (or inverse temperature) to $10^7$ for training but $10^9$ for decoding, because we have found this to perform better empirically, potentially due to the cold posterior effect~\citep{pmlr-v119-wenzel20a}.

\revision{For training with AdamW, we set $(\beta_1, \beta_2) = (0.9, 0.999)$ and perform a sweep over learning rates $\{1\cdot10^{-5}, 1\cdot10^{-4}, 5\cdot10^{-4} \}$. We again anneal the learning rates to $0$, set a small weight decay of $10^{-6}$ and rescale gradients to unit norm but do not clip them element-wise.}

We train for $1$ epoch for IWSLT17 and XSUM, 5 epochs for E2ENLG, 2 epochs for WMT18, and for 4 epochs on SamSUM.
We always take the final checkpoints after training has ended.

\subsection{Zero-shot results}
In addition to trained models, we also evaluate zero-shot prompted models.
While we do not have an explicit posterior in this setting, ensembling such models can be understood as a crude approximation to sampling from the unknown Bayes posterior.

\paragraph{Datasets}
In addition to IWSLT17 De-En and XSUM, which are described in~\cref{app:finetuning}, we use the Cs-En partition of WMT19 (\url{https://huggingface.co/datasets/wmt/wmt19})~\citep{barrault-etal-2019-findings}.
On XSUM we only evaluate on the first 1000 examples of the test set due to computational load.

\paragraph{Models}
\label{app:zero_shot}
\begin{table*}[t]
    \centering
    \resizebox{\linewidth}{!}{\begin{tabular}{ll}
         Dataset & Instruction \\
    \hline
         IWSLT17 De-En & Translate the following English text to German. Make sure to only generate the translation without extra text:  \\ 
         WMT19 Cs-En & Translate the following Czech text to English. Make sure to only generate the translation without extra text: \\ 
         XSUM & Given a BBC article, write a short summary of the article in one sentence. \\
         \hline
    \end{tabular}}
    \caption{Prompts used for zero-shot experiments.}
    \label{tab:zero_shot_prompts}
\end{table*}
We use different models for our experiments.
In particular, we use Gemma-2 9B (\url{https://huggingface.co/google/gemma-2-9b-it})~\citep{gemmateam2024gemma2improvingopen}, Llama-3 8B (\url{https://huggingface.co/meta-llama/Llama-3.1-8B-Instruct})~\citep{dubey2024llama3herdmodels}, Mistral 7B (\url{https://huggingface.co/mistralai/Mistral-7B-Instruct-v0.3})~\citep{jiang2023mistral7b}, and Qwen-2 7B (\url{https://huggingface.co/Qwen/Qwen2-7B-Instruct})~\citep{yang2024qwen2technicalreport}.
We use the instruction-tuned version of each model.
We select the models used for each dataset based on a manual inspection of their performance on each dataset.
For example, Gemma sometimes returned czech text when asked to translate from czech to english and was therefore not included in the experiment, and Mistral tended to produce too long summaries for XSUM when compared to other models.
We use the following models for each dataset:
Gemma-2, Llama-3, and Mistral for IWSLT17, Gemma-2, Qwen-2, Llama-3 for XSUM, and Llama-3 and Mistral for WMT19.
The prompts are shown in~\cref{tab:zero_shot_prompts}
Our prompt for XSUM is taken from~\citep{suzgun-etal-2023-follow}.

\paragraph{Decoding}
We use ancestral sampling with a temperature of $1.0$ for all experiments.

\subsection{Hypothesis set sizes}
\label{app:experimental_details_hyp_sizes}
For the finetuning experiments, we use 40 candidate hypotheses for the single model baseline and token-level combination, and 20 per model for~\cref{eq:seq_level_mbr_estimator} and 10 per model for~\cref{eq:hyp_set_concat_estimator}, except for XSUM, where we use 20, 10, and 5 candidate hypotheses, respectively.

\subsection{Selective Prediction}
For selective prediction we reuse the models and set-up from~\cref{app:from_scratch} which were used for~\cref{tab:main_results}.
In particular, we use the sequence-level model combination of~\cref{eq:seq_level_mbr_estimator} and token-level combination with both ancestral sampling and beam search.
The beam size is always 40 for MBR@mean, 20 for each model used in sequence-level combination and 10 for each model used in token-level combination.
All training details are the same as in~\cref{app:from_scratch}.

\subsection{Scaling experiment}
Again, we use the set-up from~\cref{app:from_scratch} with Transformer$_\text{base}$ trained from scratch on IWSLT14.
We scale all methods according to the same training recipe as described there but with different random seeds to train the different models.

\section{Additional Results}

\subsection{Results on afroMT}
\label{app:afromt}
\begin{table*}[t!]
    \centering
    \setlength{\tabcolsep}{5pt}
\def\arraystretch{1.005}
    \resizebox{\linewidth}{!}{\begin{tabular}{lcccccccccc}
       &\multicolumn{4}{c}{ AfroMT En-Bem} & \multicolumn{4}{c}{ AfroMT En-Run}& MBR & \\ 
        & \multicolumn{2}{c}{Sampling} & \multicolumn{2}{c}{Beam Search} & \multicolumn{2}{c}{Sampling} & \multicolumn{2}{c}{Beam Search} & MBR & Effective\\
       & BLEU & chrF & BLEU & chrF & BLEU & chrF & BLEU & chrF & comparisons & beam size \\ \hline
       \; MBR (Mean) & 18.26 & 47.47 & 19.70 & 49.02 & 24.97 & 53.29 & 26.67 & 54.79 & 400 & 20\\
         & 18.63 & 47.89 & 19.70 & 49.02 & 25.58 & 53.76 & 26.67 & 54.80 & 1600 & 40\\
        \multicolumn{11}{l}{\rule{0pt}{0.1in}\bf Sequence-level (\cref{eq:hyp_set_concat_estimator})} \\
        Unimodal & 18.58 & 47.84 & 19.46 & 48.88 & 25.80 & 53.86 & 26.38 & 54.65 & 1600 & 40 \\
        Deep Ensemble & {\bf 19.71} & {\bf 48.77} & 21.28 & 50.35 & {\bf 26.52} & {\bf 54.56} & 28.19 & 56.02 & 1600 & 40 \\
        \multicolumn{11}{l}{\rule{0pt}{0.1in}\bf Sequence-level (\cref{eq:seq_level_mbr_estimator})} \\
       \; Unimodal & 18.43 & 47.75 & 19.62 & 48.95 & 25.34 & 53.66 & 26.58 & 54.77 & 1600 & 80\\
       \; Deep Ensemble & 19.48 & 48.49& 20.69 & 49.88 & 25.86 & 54.22 & 27.40 & 55.42 & 1600 & 80\\
        \multicolumn{11}{l}{\rule{0pt}{0.1in}\bf Token-level} \\
       \; Unimodal & 17.90 & 47.29 & 19.60 & 48.94 & 24.86 & 53.29 & 26.57 & 54.79 & 400 & 80\\
       \; Deep Ensemble & 19.32 & 48.49& {\bf 21.51} & {\bf 50.54} & 25.46 & 53.71 & {\bf 28.44} & {\bf 56.28} & 400 & 80 \\ 
       \hline
    \end{tabular}}
    \caption{
Results on afroMT with Transformer$_\text{base}$ trained from scratch.
    \label{tab:afromt}
    }
\end{table*}
\cref{tab:afromt} shows results on the En-Run and En-Bem partitions of afroMT.
We find similar patterns to our results presented in~\cref{tab:main_results}: Deep-Ensemble-based weight uncertainty always improves performance, even with matched compute budgets, while unimodal posteriors perform similarly to a single model baseline.

\subsection{Results with LaBSE for from-scratch-trained models}
\begin{table*}[t]
    \centering
    \resizebox{.7\textwidth}{!}{
    \begin{tabular}{lcccccc}
        &  \multicolumn{3}{c}{Sampling} & \multicolumn{3}{c}{Beam Search} \\
         Method & BLEU & COMET & LaBSE & BLEU & COMET & LaBSE \\
         \hline
        \; MBR@Mean & 33.69 & 74.71 & 85.33 & 35.90 & 76.65 & 86.44 \\
        \multicolumn{7}{l}{\rule{0pt}{0.1in}\bf Sequence-level - \cref{eq:hyp_set_concat_estimator}} \\
        Unimodal & 34.59 & 75.15 & 85.65 & 35.78 & 76.55& 86.42 \\
        Deep Ensemble & {\bf 36.03} & 75.79 & 85.98 & 38.30 & 78.01 & 87.16 \\
        \multicolumn{7}{l}{\rule{0pt}{0.1in}\bf Sequence-level - \cref{eq:seq_level_mbr_estimator}} \\
        \; Unimodal & 34.65 & 75.20 & 85.68 & 35.99 & 76.67 & 86.45\\
        \; Mixture & 35.42 & {\bf 75.84} & {\bf 86.07} & 37.42 & 77.69 & 86.97 \\
        \multicolumn{7}{l}{\rule{0pt}{0.1in}\bf Token-level} \\
        \; Unimodal & 33.62 & 74.68 & 85.39 & 35.94 & 76.66 & 86.45 \\
        \; Mixture & 34.61 & 75.06 & 85.88 & {\bf 38.56} & {\bf 78.31} & {\bf 87.34} \\ \hline
    \end{tabular}
    }
    \caption{
    Measuring hallucinations with LaBSE (higher is better) on IWSLT14 with Transformer$_\text{base}$ shows similar trends as quality estimation metrics: incorporating weight-uncertainty can reduce hallucinations, especially when a complex posterior is used. Here, we use a hypothesis set size of 20 for all methods but~\cref{eq:hyp_set_concat_estimator} which uses a size of 10.
    }
    \label{tab:hallucinations_iwslt}
\end{table*}

\begin{table*}[t]
    \centering
    \resizebox{.7\textwidth}{!}{
    \begin{tabular}{lcccccc}
        &  \multicolumn{3}{c}{Sampling} & \multicolumn{3}{c}{Beam Search} \\
         Method & BLEU & COMET & LaBSE & BLEU & COMET & LaBSE \\
         \hline
        \; MBR@Mean & 23.37 & 71.04 & 86.97 & 27.56 & 75.23 & 88.46 \\
        \multicolumn{7}{l}{\rule{0pt}{0.1in}\bf Sequence-level - \cref{eq:hyp_set_concat_estimator}} \\
        Unimodal & 24.31 & 72.09 & 87.36 & 27.52 & 75.16 & 88.42 \\
        Deep Ensemble & {\bf 24.70} & 72.39 & {\bf 87.61} & {\bf 28.99} & 76.02
 & 88.68 \\
        \multicolumn{7}{l}{\rule{0pt}{0.1in}\bf Sequence-level - \cref{eq:seq_level_mbr_estimator}} \\
        Unimodal & 24.21 & 72.15 & 87.32 & 27.56 & 75.21 & 88.44 \\
        Deep Ensemble & 24.67 & {\bf 72.58} & 87.56 & 28.29 & 75.70
 & 88.75 \\
        \multicolumn{7}{l}{\rule{0pt}{0.1in}\bf Token-level} \\
        Unimodal & 23.44 & 71.36 & 86.84 & 27.75 & 75.19 & 88.35 \\
        Deep Ensemble & 23.95 & 71.58 & 87.16 & 28.98 & {\bf 76.08} & 88.75 \\
 \hline 
 \end{tabular}
    }
    \caption{
    Measuring hallucinations with LaBSE (higher is better) on WMT14 with Transformer$_\text{large}$ shows similar trends as quality estimation metrics: incorporating weight-uncertainty can reduce hallucinations, especially when a complex posterior is used. Here, we use a hypothesis set size of 20 for all methods but~\cref{eq:hyp_set_concat_estimator} which uses a size of 10.
    }
    \label{tab:hallucinations_wmt}
\end{table*}
\cref{tab:hallucinations_iwslt} and \cref{tab:hallucinations_wmt} show LaBSE scores for hallucination evaluation for the same evaluation setting as in~\cref{tab:main_results}.
Again, we find hallucinations to be reduced when weight uncertainty is accounted for.

\begin{table}[t!]
    \centering
        \resizebox{.9\linewidth}{!}{\begin{tabular}{lccccc}
        Method & Creation of $\hypset$ (s) & Utility Calculation (s) & Total (s) & R-1 & R-L\\ \hline
        MBR@mean & 5824 & 402 & 6226 & 68.74 & 45.16 \\
        Sequence-level~\cref{eq:hyp_set_concat_estimator} & 5472 & 408 & 5880 & 69.36 & 45.57\\
        Sequence-level~\cref{eq:seq_level_mbr_estimator} & 5881 & 418 & 6299 & 69.13 & 45.38\\
        \hline
    \end{tabular}}
    \caption{\revision{Time (in seconds) taken for decoding for the results on E2ENLG from~\cref{tab:llm}.}}
    \label{tab:inference_time}
\end{table}
\subsection{\revision{Inference-Time Measurements}}
\label{app:decoding_time}
\revision{\cref{tab:inference_time} shows the time needed for decoding in seconds as well as the obtained results for the E2ENLG experiment from~\cref{tab:llm}.
All results were obtained on NVIDIA GeForce RTX 3090 GPUs with 24GB memory.}

\end{document}